\documentclass[11pt]{article}
\usepackage[a4paper,top=2.5cm,bottom=2.5cm,left=2.5cm,right=2.5cm,marginparwidth=1.75cm]{geometry}

\usepackage{natbib}
\setcitestyle{numbers,square,comma}
\usepackage{times}
\usepackage{soul}
\usepackage{url}
\usepackage[utf8]{inputenc}
\usepackage{graphicx}
\usepackage{amsmath}
\usepackage{booktabs}
\usepackage[linesnumbered, ruled, vlined]{algorithm2e}
\usepackage{float}
\usepackage[caption=false]{subfig}

\usepackage{amsmath,amsfonts,amssymb}
\usepackage{adjustbox}
\usepackage{multirow}
\usepackage{color}

\newcommand{\del}[1]{}

\usepackage{microtype}
\usepackage{booktabs} % for professional tables
\usepackage{diagbox}
\usepackage{pifont}
\newcommand{\cmark}{\ding{51}}
\newcommand{\xmark}{\ding{55}}

\usepackage{makecell}
\usepackage{arydshln}

\usepackage{amsthm}
\usepackage{amssymb}
\theoremstyle{definition}

\graphicspath{figure}
\usepackage{xcolor}
\usepackage{multirow}

\usepackage{subfiles}
\usepackage{wrapfig}
\usepackage{authblk}

\newcommand{\rev}[1]{{\textcolor{black}{#1}}}
\newcommand{\SXH}[1]{{\textcolor{black}{#1}}}
\newcommand{\QB}[1]{{\textcolor{black}{#1}}}
\newcommand{\QBB}[1]{{\textcolor{black}{#1}}}
\newcommand{\tb}[1]{\textbf{#1}}

\newcommand{\LY}[1]{{\textcolor{black}{#1}}}
\newcommand{\LX}[1]{{\textcolor{black}{#1}}}

\newcommand{\NB}[1]{{\textcolor{black}{#1}}}
\newcommand{\NBP}[1]{{\textcolor{black}{#1}}}

\newcommand{\REF}[1]{{\textcolor{black}{#1}}}

\newcommand{\HBS}[1]{{\textcolor{black}{#1}}}

\date{}

\begin{document}

\title{A Survey on Federated Learning Systems: Vision, Hype and Reality for Data Privacy and Protection}

\author{
Qinbin Li$^{1}$,
Zeyi Wen$^2$,
Zhaomin Wu$^1$,
Sixu Hu$^1$,

Naibo Wang$^1$,
Yuan Li$^{1}$,
Xu Liu$^{1}$,
Bingsheng He$^1$\\
$^{1}$National University of Singapore\\
$^2$The University of Western Australia\\
$^1$\{qinbin, zhaomin, sixuhu, naibowang, liyuan, liuxu, hebs\}@comp.nus.edu.sg\\
$^2$zeyi.wen@uwa.edu.au\\
}

\maketitle

\begin{abstract}
Federated learning has been a hot research topic in enabling the collaborative training of machine learning models among different organizations under the privacy restrictions. As researchers try to support more machine learning models with different privacy-preserving approaches, there is a requirement in developing systems and infrastructures to ease the development of various federated learning algorithms. Similar to deep learning systems such as PyTorch and TensorFlow that boost the development of deep learning, federated learning systems (FLSs) are equivalently important, and face challenges from various aspects such as effectiveness, efficiency, and privacy. In this survey, we conduct a comprehensive review on federated learning systems. To achieve smooth flow and guide future research, we introduce the definition of federated learning systems and analyze the system components. Moreover, we provide a thorough categorization for federated learning systems according to six different aspects, including data distribution, machine learning model, privacy mechanism, communication architecture, scale of federation and motivation of federation. The categorization can help the design of federated learning systems as shown in our case studies. By systematically summarizing the existing federated learning systems, we present the design factors, case studies, and future research opportunities.
\end{abstract}

\section{Introduction}
Many machine learning algorithms are data hungry, and in reality, data are dispersed over different organizations under the protection of privacy restrictions. Due to these factors, federated learning (FL)~\cite{mcmahan2016communication,yang2019federated,kairouz2019advances} has become a hot research topic in machine learning. For example, data of different hospitals are isolated and become ``data islands''. Since each data island has limitations in size and approximating real distributions, a single hospital may not be able to train a high-quality model that has a good predictive accuracy for a specific task. Ideally, hospitals can benefit more if they can collaboratively train a machine learning model on the union of their data. However, the data cannot simply be shared among the hospitals due to various policies and regulations. Such phenomena on ``data islands'' are commonly seen in many areas such as finance, government, and supply chains. Policies such as General Data Protection Regulation (GDPR)~\cite{albrecht2016gdpr}\del{ and Personal Information Security Specification (PISS)~\cite{berger2019national}} stipulate rules on data sharing among different organizations. Thus, it is challenging to develop a federated learning system which has a good predictive accuracy while obeying policies and regulations to protect privacy.

% todo: simplify the paragraph
\HBS{Many efforts have recently been devoted to implementing federated learning algorithms to support effective machine learning models. Specifically, researchers try to support more machine learning models with different privacy-preserving approaches, including deep neural networks (NNs)~\cite{liu2018secure,yurochkin2019bayesian,bonawitz2019towards,ryffel2018generic,mcmahan2016communication}, gradient boosted decision trees (GBDTs)~\cite{zhao2018inprivate,cheng2019secureboost,li2019practical}, logistics regression~\cite{nikolaenko2013privacy,chen2018privacy} and support vector machines (SVMs)~\cite{smith2017federated}.} For instance, \citet{nikolaenko2013privacy} and \citet{chen2018privacy} propose approaches to conduct FL based on linear regression. Since GBDTs have become very successful in recent years~\cite{chen2016xgboost,ThunderGBM}, the corresponding Federated Learning Systems (FLSs) have also been proposed by \citet{zhao2018inprivate}, \citet{cheng2019secureboost},~\citet{li2019practical}. Moreover, there are many FLSs supporting the training of NNs. Google proposes a scalable production system which enables tens of millions of devices to train a deep neural network~\cite{bonawitz2019towards}.

\HBS{As there are common methods and building blocks (e.g., privacy mechanisms such as differential privacy) for building FL algorithms, it makes sense to develop systems and infrastructures to ease the development of various FL algorithms. Systems and infrastructures allow algorithm developers to reuse the common building blocks, and avoid building algorithms every time from scratch. Similar to deep learning systems such as PyTorch~\cite{paszke2017automatic,paszke2019pytorch} and TensorFlow~\cite{abadi2016tensorflow} that boost the development of deep learning algorithms, FLSs are equivalently important for the success of FL. However, building a successful FLS is challenging, which needs to consider multiple aspects such as effectiveness, efficiency, privacy, and autonomy.}

In this paper, we take a survey on the existing FLSs from a system view. First, we show the definition of FLSs, and compare it with conventional federated systems. Second, we analyze the system components of FLSs, including the parties, the manager, and the computation-communication framework. Third, we categorize FLSs based on six different aspects: data distribution, machine learning model, privacy mechanism, communication architecture, scale of federation, and motivation of federation. These aspects can direct the design of an FLS as common building blocks and system abstractions. Fourth, based on these aspects, we systematically summarize the existing studies, which can be used to direct the design of FLSs. Last, to make FL more practical and powerful, we present future research directions to work on. \HBS{We believe that systems and infrastructures are essential for the success of FL.  More work has to be carried out to address the system research issues in effectiveness, efficiency, privacy, and autonomy.}

\subsection{Related Surveys}
\REF{There have been several surveys on FL. A seminal survey written by Yang et al.~\cite{yang2019federated} introduces the basics and concepts in FL, and further proposes a comprehensive secure FL framework. The paper mainly target at a relatively small number of parties which are typically enterprise data owners. \citet{li2019federated} summarize challenges and future directions of FL in massive networks of mobile and edge devices. Recently, \citet{kairouz2019advances} have a comprehensive description about the characteristics and challenges on FL from different research topics. However, they mainly focus on cross-device FL, where the participants are a very large number of mobile or IoT devices. More recently, another survey \cite{aledhari2020federated} summarizes the platforms, protocols and applications of federated learning. Some surveys only focus on an aspect of federated learning. For example, \citet{lim2019federated} conduct a survey of FL specific to mobile edge computing, while \cite{lyu2020threats} focuses on the threats to federated learning.}

\subsection{Our Contribution}
To the best of our knowledge, there lacks a survey on reviewing existing systems and infrastructure of FLSs and on boosting the attention of creating systems for FL (Similar to prosperous system research in deep learning). \REF{In comparison with the previous surveys,} the main contributions of this paper are as follows. \QB{(1) Our survey is the first one to provide a comprehensive analysis on FL from a system's point of view, including system components, taxonomy, summary, design, and vision.} (2) We provide a comprehensive taxonomy against FLSs on six different aspects, including data distribution, machine learning model, privacy mechanism, communication architecture, scale of federation, and motivation of federation, which can be used as common building blocks and system abstractions of FLSs. (3) We summarize existing typical and state-of-the-art studies according to their domains, which is convenient for researchers and developers to refer to. (4) We present the design factors for a successful FLS and comprehensively review solutions for each scenario. (5) We propose interesting research directions and challenges for future generations of FLSs.

The rest of the paper is organized as follows. In Section \ref{sec:fedsys}, we introduce the concept and the system components of FLSs. In Section \ref{sec:taxonomy}, we propose six aspects to classify FLSs. In Section \ref{sec:exist_study}, we summary existing studies and systems on FL. We then present the design factors and solutions for an FLS in Section \ref{sec:design}. Last, we propose possible future directions on FL in Section \ref{sec:vision} and conclude our paper in Section \ref{sec:conc}.

\section{An Overview of Federated Learning Systems}
\label{sec:fedsys}

\subsection{Background}

As data breach becomes a major concern, more and more governments establish regulations to protect users' data, such as GDPR in European Union~\cite{voigt2017eu}, PDPA in Singapore~\cite{chik2013singapore}, and CCPA~\cite{ccpa_link} in the US. The cost of breaching these policies is pretty high for companies. In a breach of 600,000 drivers' personal information in 2016, Uber had to pay \$148 million to settle the investigation~\cite{uberbreach2019}. SingHealth was fined \$750,000 by the Singapore government for a breach of PDPA~\cite{singbreach2019}. Google was fined \$57 million for a breach of GDPR~\cite{googlebreach2019}, which is the largest penalty as of March 18, 2020 under the European Union privacy law.

Under the above circumstances, federated learning, a collaborative learning without exchanging users' original data, has drawn increasingly attention nowadays. While machine learning, especially deep learning, has attracted many attentions again recently, the combination of federation and machine learning is emerging as a new and hot research topic.

\subsection{Definition}
FL enables multiple parties jointly train a machine learning model without exchanging the local data. It covers the techniques from multiple research areas such as distributed system, machine learning, and privacy. \rev{Inspired by the definition of FL given by other studies \cite{kairouz2019advances,yang2019federated}, here we give a definition of FLSs.}

\rev{In a federated learning system, multiple parties collaboratively train machine learning models without exchanging their raw data. The output of the system is a machine learning model for each party (which can be same or different). A practical federated learning system has the following constraint: given an evaluation metric such as test accuracy, the performance of the model learned by federated learning should be better than the model learned by local training with the same model architecture.}

\subsection{Compare with Conventional Federated Systems}
\label{sec:comp_fed}
The concept of federation can be found with its counterparts in the real world such as business and sports. \del{For example, the United States is a federation of 50 self-governing states. Malaysia is also a federation of Sarawak and Sabah.} The main characteristic of federation is cooperation. Federation not only commonly appears in society, but also plays an important role in computing. In computer science, federated computing systems have been an attractive area of research under different contexts.

Around 1990, there were many studies on federated database systems (FDBSs)~\cite{sheth1990federated}. An FDBS is a collection of autonomous databases cooperating for mutual benefits. As pointed out in a previous study~\cite{sheth1990federated}, three important components of an FDBS are autonomy, heterogeneity, and distribution. 
\begin{itemize}
    \item \emph{Autonomy}. A database system (DBS) that participates in an FDBS is autonomous, which means it is under separate and independent control. The parties can still manage the data without the FDBS.
    \item \emph{Heterogeneity}. The database management systems can be different inside an FDBS. For example, the difference can lie in the data structures, query languages, system software requirements, and communication capabilities.
    \item \emph{Distribution}. Due to the existence of multiple DBSs before an FDBS is built, the data distribution may differ in different DBSs. A data record can be horizontally or vertically partitioned into different DBSs, and can also be duplicated in multiple DBSs to increase the reliability.
\end{itemize}

More recently, with the development of cloud computing, many studies have been done for federated cloud computing~\cite{kurze2011cloud}. A federated cloud (FC) is the deployment and management of multiple external and internal cloud computing services. The concept of cloud federation enables further reduction of costs due to partial outsourcing to more cost-efficient regions. Resource migration and resource redundancy are two basic features of federated clouds~\cite{kurze2011cloud}. First, resources may be transferred from one cloud provider to another. Migration enables the relocation of resources. Second, redundancy allows concurrent usage of similar service features in different domains. For example, the data can be partitioned and processed at different providers following the same computation logic. Overall, the scheduling of different resources is a key factor in the design of a federated cloud system.

There are some similarities and differences between FLSs and conventional federated systems. First, the concept of federation still applies. The common and basic idea is about the cooperation of multiple independent parties. Therefore, the perspective of considering heterogeneity and autonomy among the parties can still be applied to FLSs. Second, some factors in the design of distributed systems are still important for FLSs. For example, how the data are shared between the parties can influence the efficiency of the systems. For the differences, these federated systems have different emphasis on collaboration and constraints. While FDBSs focus on the management of distributed data and FCs focus on the scheduling of the resources, FLSs care more about the secure computation among multiple parties. FLSs induce new challenges such as the algorithm designs of the distributed training and the data protection under the privacy restrictions.

\begin{figure}
\begin{center}
\includegraphics[width=0.8\columnwidth]{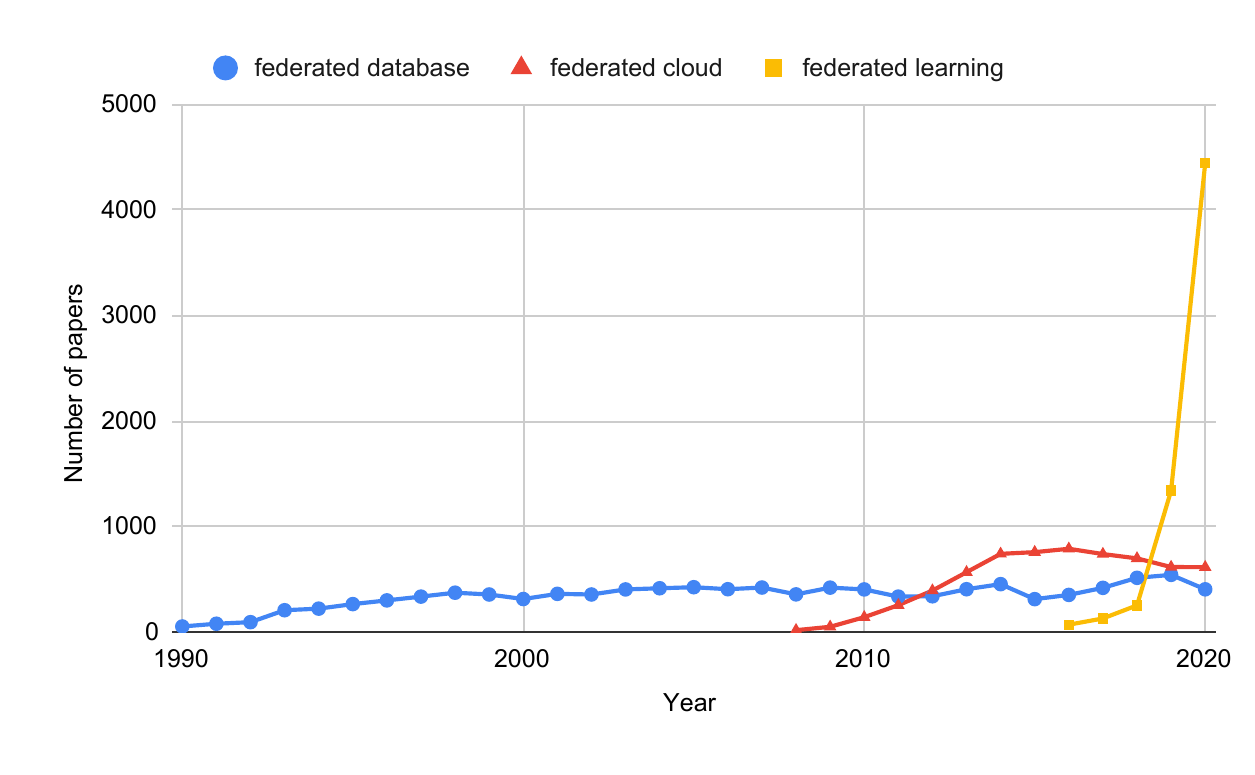}
\caption{The number of related papers on ``federated database'', ``federated cloud'', and ``federated learning''}
\label{fig:trend}
\end{center}
\end{figure}

Figure~\ref{fig:trend} shows the number of papers in each year for these three research areas. Here we count the papers by searching keywords ``federated database'', ``federated cloud'', and ``federated learning'' in Google Scholar\footnote{\url{https://scholar.google.com/}}. Although federated database was proposed 30 years ago, there are still about 400 papers that mentioned it in recent years. The popularity of federated cloud grows more quickly than federated database at the beginning, while it appears to decrease in recent years probably because cloud computing becomes more mature and the incentives of federation diminish. For FL, the number of related papers is increasing rapidly and has achieved about 4,400 last year. Nowadays, the ``data island'' phenomena are common and have increasingly become an important issue in machine learning. Also, there is a increasing privacy concern and social awareness from the general public. Thus, we expect the popularity of FL will keep increasing for at least five years until there may be mature FLSs.

\subsection{System Components}
\label{sec:sys_com}
There are three major components in an FLS: parties (e.g., clients), the manager (e.g., server), and the communication-computation framework to train the machine learning model.

\subsubsection{Parties}
In FLSs, the parties are the data owners and the beneficiaries of FL. They can be organizations or mobile devices, named cross-silo or cross-device settings~\cite{kairouz2019advances}, respectively. We consider the following properties of the parties that affect the design of FLSs.

First, what is the hardware capacity of the parties? The hardware capacity includes the computation power and storage. If the parties are mobile phones, the capacity is weak and the parties cannot perform much computation and train a large model. For example, \citet{wang2019adaptive} consider a resource constrained setting in FL. They design an objective to include the resource budget and proposed an algorithm to determine the rounds of local updates.

Second, what is the scale and stability of the parties? For organizations, the scale is relative small compared with the mobile devices. Also, the stability of the cross-silo setting is better than the cross-device setting. Thus, in the cross-silo setting, we can expect that every party can continuously conduct computation and communication tasks in the entire federated process, which is a common setting in many studies~\cite{li2019practical,cheng2019secureboost,smith2017federated}. If the parties are mobile devices, the system has to handle possible issues such as connection lost~\cite{bonawitz2019towards}. Moreover, since the number of devices can be very large (e.g., millions), it is unpractical to assume all the devices to participate every round in FL. The widely used setting is to choose a fraction of devices to perform computation in each round~\cite{mcmahan2016communication,bonawitz2019towards}.

Last, what are the data distributions among the parties? Usually, no matter cross-device or cross-silo setting, the non-IID (identically and independently distributed) data distribution is considered a practical and challenging setting in federated learning~\cite{kairouz2019advances}, which is evaluated in the experiments of recent work~\cite{li2019practical,yurochkin2019bayesian,li2019convergence,wang2020federated}. Such non-IID data distribution may be more obvious among the organizations. For example, a bank and an insurance company can conduct FL to improve their predictions (e.g., whether a person can repay the loan and whether the person will buy the insurance products), while even the features can vary a lot in these organizations. Techniques in transfer learning~\cite{pan2010survey}, meta-learning~\cite{finn2017model}, and multi-task learning~\cite{ruder2017overview} may be useful to combine the knowledge of various kinds of parties.

\subsubsection{Manager}
In the cross-device setting, the manager is usually a powerful central server. It conducts the training of the global machine learning model and manages the communication between the parties and the server. The stability and reliability of the server are quite important. Once the server fails to provide the accurate computation results, the FLS may produce a bad model. To address these potential issues, blockchain~\cite{swan2015blockchain} may be a possible technique to offer a decentralized solution in order to increase the system reliability. For example,~\citet{kim2018device} leverage the blockchain in lieu of the central server in their system, where the blockchain enables exchanging the devices' updates and providing rewards to them.

In the cross-silo setting, since the organizations are expected to have powerful machines, the manager can also be one of the organizations who dominates the FL process. This is particularly used in the vertical FL~\cite{yang2019federated}, which we will introduce in Section~\ref{sec:data_distribution} in detail. In a vertical FL setting by~\citet{liu2018secure}, the features of data are vertically partitioned across the parties and only one party has the labels. The party that owns the labels is naturally considered as the FL manager.

One challenge can be that it is hard to find a trusted server or party as the manager, especially in the cross-silo setting. Then, a fully-decentralized setting can be a good choice, where the parties communicate with each other directly and almost equally contribute to the global machine learning model training. These parties jointly set a FL task and deploy the FLS.~\citet{li2019practical} propose a federated gradient boosting decision trees framework, where each party trains decision trees sequentially and the final model is the combination of all trees. It is challenging to design a fully-decentralized FLS with reasonable communication overhead.

% SGX

\subsubsection{Communication-Computation Framework}
\label{sec:framework}
In FLSs, the computation happens on the parties and the manager, while the communication happens between the parties and the manager. Usually, the aim of the computation is for the model training and the aim of the communication is for exchanging the model parameters.

\begin{figure}
% \captionsetup[subfloat]{farskip=2pt,captionskip=1pt}
\centering
\subfloat[FedAvg]{\includegraphics[width=.45\columnwidth]{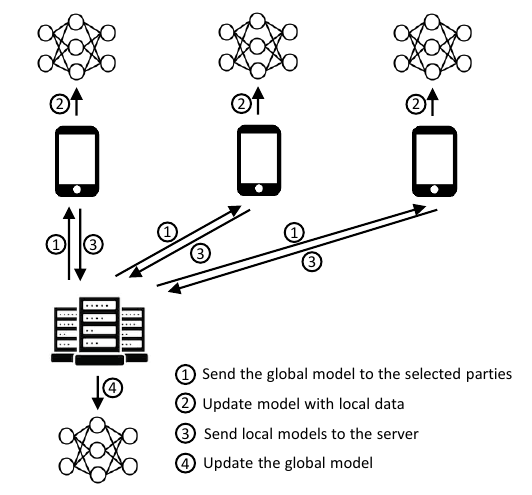}%
\label{fig:sgd_frame}
}
\subfloat[SimFL]{\includegraphics[width=.45\columnwidth]{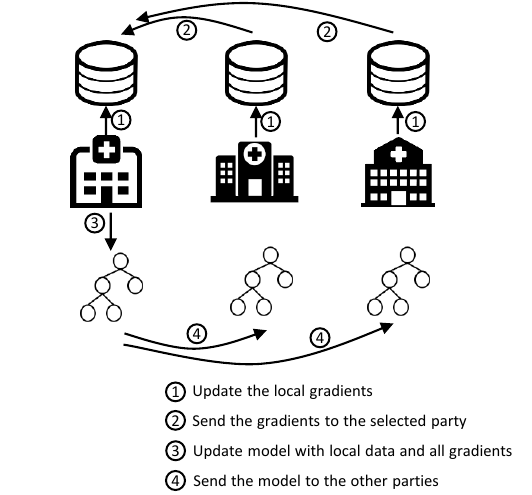}%
\label{fig:decen_frame}
}
\caption{Federated learning frameworks}
\label{fig:frameworks}
\end{figure}

A basic and widely used framework is Federated Averaging (FedAvg)~\cite{mcmahan2016communication} proposed in 2016, as shown in Figure~\ref{fig:sgd_frame}. In each iteration, the server first sends the current global model to the selected parties. Then, the selected parties update the global model with their local data. Next, the updated models are sent back to the server. Last, the server averages all the received local models to get a new global model. FedAvg repeats the above process until reaching the specified number of iterations. The global model of the server is the final output.

While FedAvg is a centralized FL framework, SimFL, proposed by \citet{li2019federated}, represents a decentralized FL framework. In SimFL, no trusted server is needed. In each iteration, the parties first update the gradients of their local data. Then, the gradients are sent to a selected party. Next, the selected party use its local data and the gradients to update the model. Last, the model is sent to all the other parties. To ensure fairness and utilize the data from different parties, every party is selected for updating the model for about the same number of rounds. SimFL repeats a specified number of iterations and outputs the final model.

\section{Taxonomy}
\label{sec:taxonomy}
\begin{figure*}
\begin{center}
\includegraphics[width=\textwidth]{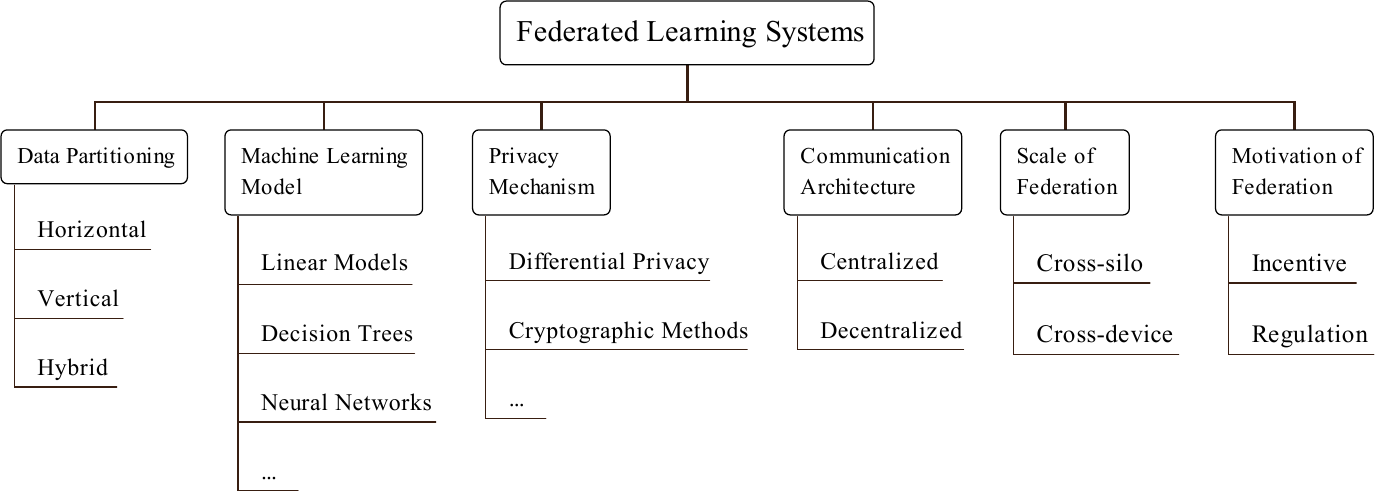}
\caption{Taxonomy of federated learning systems}
\label{fig:taxonomy}
\end{center}
\end{figure*}

Considering the common system abstractions and building blocks for different FLSs, we classify FLSs by six aspects: data partitioning, machine learning model, privacy mechanism, communication architecture, scale of federation, and motivation of federation. These aspects include common factors (e.g., data partitioning, communication architecture) in previous FLSs \cite{sheth1990federated,kurze2011cloud} and unique consideration (e.g., machine learning model and privacy mechanism) for FLSs. Furthermore, these aspects can be used to guide the design of FLSs. Figure~\ref{fig:taxonomy} shows the summary of the taxonomy of FLSs.

\rev{In Table 1 of \cite{kairouz2019advances}, they consider different characteristics to distinguish distributed learning, cross-device federated learning, and cross-silo federated learning, including setting, data distribution, communication, etc. Our taxonomy is used to distinguish different federated learning systems from a deployment view, and aspects like machine learning models and motivation of federation are not considered in \cite{kairouz2019advances}.}

\subsection{Data Partitioning}
\label{sec:data_distribution}
Based on how data are distributed over the sample and feature spaces, FLSs can be typically categorized in horizontal, vertical, and hybrid FLSs~\cite{yang2019federated}. 

In horizontal FL, the datasets of different parties have the same feature space but little intersection on the sample space. This is a natural data partitioning especially for the cross-device setting, where different users try to improve their model performance on the same task using FL. Also, the majority of FL studies adopt horizontal partitioning. Since the local data are in the same feature space, the parties can train the local models using their local data with the same model architecture. The global model can simply be updated by averaging all the local models. A basic and popular framework of horizontal federated learning is FedAvg, as shown in Figure \ref{fig:frameworks}. Wake-word recognition~\cite{leroy2019federated}, such as `Hey Siri' and `OK Google', is a typical application of horizontal partition because each user speaks the same sentence with a different voice.

In vertical FL, the datasets of different parties have the same or similar sample space but differ in the feature space. For the vertical FLS, it usually adopts \textit{entity alignment} techniques \cite{yan2016survey,christen2012data} to collect the overlapped samples of the parties. Then the overlapped data are used to train the machine learning model using encryption methods. \citet{cheng2019secureboost} propose a lossless vertical FLS to enable parties to collaboratively train gradient boosting decision trees. They use privacy-preserving entity alignment to find common users among two parties, whose gradients are used to jointly train the decision trees. Cooperation among different companies usually can be treated as a situation of vertical partition. 

In many other applications, while existing FLSs mostly focus on one kind of partition, the partition of data among the parties may be a hybrid of horizontal partition and vertical partition. Let us take cancer diagnosis system as an example. A group of hospitals wants to build an FLS for cancer diagnosis but each hospital has different patients as well as different kinds of medical examination results. Transfer learning~\cite{pan2010survey} is a possible solution for such scenarios. \citet{liu2018secure} propose a secure federated transfer learning system which can learn a representation among the features of parties using common instances.

\subsection{Machine Learning Models}

Since FL is used to solve machine learning problems, the parties usually want to train a state-of-the-art machine learning model on a specified task. There have been many efforts in developing new models or reinventing current models to the federated setting. Here, we consider the widely-used models nowadays. The most popular machine learning model now is neural network (NN), which achieves state-of-the-art results in many tasks such as image classification and word prediction~\cite{krizhevsky2012imagenet,sundermeyer2012lstm}. There are many federated learning studies based on stochastic gradient descent~\cite{mcmahan2016communication,wang2020federated,bonawitz2019towards}, which can be used to train NNs.

Another widely used model is decision tree, which is highly efficient to train and easy to interpret compared with NNs. A tree-based FLS is designed for the federated training of single or multiple decision trees (e.g., gradient boosting decision trees (GBDTs) and random forests). GBDTs are especially popular recently and it has a very good performance in many classification and regression tasks\del{~\cite{chen2016xgboost}}. \citet{li2019practical} and \citet{cheng2019secureboost} propose FLSs for GBDTs on horizontally and vertically partitioned data, respectively.

Besides NNs and trees, linear models (e.g., linear regression, logistic regression, SVM) are classic and easy-to-use models. There are some well developed systems for linear regression and logistic regression~\cite{nikolaenko2013privacy,hardy2017private}. These linear models are easy to learn compared with other complex models (e.g., NNs).

\rev{While a single machine learning model may be weak, ensemble methods \cite{polikar2012ensemble} such as stacking and voting can be applied in the federated setting. Each party trains a local model and sends it to the server, which aggregates all the models as an ensemble. The ensemble can directly be used for prediction by max voting or be used to train a meta-model by stacking. A benefit of federated ensemble learning is that each party can train heterogeneous models as there is no averaging of model parameters. As shown in previous studies \cite{yurochkin2019bayesian,fedkt}, federated ensemble learning can also achieve a good accuracy in a single communication round.}

Currently, many FL frameworks \cite{mcmahan2016communication,konevcny2016federated2,wang2019adaptive,li2018federated} are proposed based on stochastic gradient descent, which is a typical optimization algorithm for many models including neural networks and logistic regression. However, to increase the effectiveness of FL, we may have to exploit the model architecture \cite{wang2020federated}. Since the research of FL is still at an early stage, there is still a gap for FLSs to better support the state-of-the-art models.

\subsection{Privacy Mechanisms}

Although the local data are not exposed in FL, the exchanged model parameters may still leak sensitive information about the data. There have been many attacks against machine learning models~\cite{fredrikson2015model,shokri2017membership,nasr2019comprehensive,melis2019exploiting}, such as model inversion attack~\cite{fredrikson2015model} and membership inference attack~\cite{shokri2017membership}, which can potentially infer the raw data by accessing to the model. Moreover, there are many privacy mechanisms such as differential privacy~\cite{dwork2006calibrating} and $k$-anonymity~\cite{el2008protecting}, which provide different privacy guarantees. The characteristics of existing privacy mechanisms are summarized in the survey~\cite{wagner2018technical}. Here we introduce two major approaches that are adopted in the current FLSs for data protection: cryptographic methods and differential privacy.

Cryptographic methods such as homomorphic encryption~\REF{~\cite{aono2018privacy,hardy2017private,chabanne2017privacy,riazi2018chameleon,liu2017oblivious}}, and secure multi-party computation (SMC)~\cite{shamir1979share,chaum1988dining,bonawitz2016practical,bonawitz2017practical} are widely used in privacy-preserving machine learning algorithms. Basically, the parties have to encrypt their messages before sending, operate on the encrypted messages, and decrypt the encrypted output to get the final result. Applying the above methods, the user privacy of FLSs can usually be well protected~\cite{kantarcioglu2004privacy,yu2006privacyH,karr2009privacy,nock2018entity}. \rev{For example, SMC~\cite{goldreich1998secure} guarantees that all the parties cannot learn anything except the output, which can be used to securely aggregate the transferred gradients. However, SMC does not provide privacy guarantees for the final model, which is still vulnerable to the inference attacks and model inversion attacks \cite{shokri2017membership,fredrikson2015model}.} Also, due to the additional encryption and decryption operations, such systems suffer from the extremely high computation overhead.

Differential privacy~\cite{dwork2006calibrating,dwork2014algorithmic} guarantees that one single record does not influence much on the output of a function. Many studies adopt differential privacy~\cite{chaudhuri2011differentially,abadi2016deep,li2019privacy,thakkar2019differentially,zhao2020local,lianto2020responsive} for data privacy protection to ensure the parties cannot know whether an individual record participates in the learning or not. By injecting random noises to the data or model parameters~\cite{abadi2016deep,li2019privacy,song2013stochastic,xiao2010differential}, differential privacy provides statistical privacy guarantees for individual records and protection against the inference attack on the model. Due to the noises in the learning process, such systems tend to produce less accurate models.

\rev{Note that the above methods are independent of each other, and an FLS can adopt multiple methods to enhance the privacy guarantees~\cite{goryczka2015comprehensive,xu2019hybridalpha,2021nature}.} There are also other approaches to protect the user privacy. An interesting hardware-based approach is to use trusted execution environment (TEE) such as Intel SGX processors~\cite{sabt2015trusted,ohrimenko2016oblivious}, which can guarantee that the code and data loaded inside are protected. Such environment can be used inside the central server to increase its credibility.

Related to privacy level, the threat models also vary in FLSs \cite{lyu2020threats}. The attacks can come from any stage of the process of FL, including inputs, the learning process, and the learnt model. 
\begin{itemize}
    \item \emph{Inputs} The malicious parties can conduct data poisoning attacks \cite{chen2017targeted,li2016data,alfeld2016data} on FL. For example, the parties can modify the labels of training samples with a specific class, so that the final model performs badly on this class.
    \item \emph{Learning process} During the learning process, the parties can perform model poisoning attacks \cite{bagdasaryan2020backdoor,xie2019dba} to upload designed model parameters. Like data poisoning attacks, the global model can have a very low accuracy due to the poisoned local updating. Besides model poisoning attacks, the Byzantine fault \cite{castro1999practical,blanchard2017machine,su2018securing} is also a common issue in distributed learning, where the parties may behave arbitrarily badly and upload random updates.
    \item \emph{The learnt model}. If the learnt model is published, the inference attacks \cite{fredrikson2015model,shokri2017membership,melis2019exploiting,nasr2019comprehensive} can be conducted on it. The server can infer sensitive information about the training data from the exchanged model parameters. For example, membership inference attacks \cite{shokri2017membership,nasr2019comprehensive} can infer whether a specific data record is used in the training. Note that the inference attacks may also be conducted in the learning process by the FL manager, who has access to the local updates of the parties.
\end{itemize} 

\subsection{Communication Architecture}
There are two major ways of communication in FLSs: centralized design and decentralized design.\del{ Figure~\ref{fig:commu} shows the examples of centralized and decentralized designs.} In the centralized design, the data flow is often asymmetric, which means the manager aggregates the information (e.g., local models) from parties and sends back training results~\cite{bonawitz2019towards}. The parameter updates on the global model are always done in this manager. The communication between the manager and the local parties can be synchronous~\cite{mcmahan2016communication} or asynchronous~\cite{xie2019asynchronous,sprague2018asynchronous}. In a decentralized design, the communications are performed among the parties~\cite{zhao2018inprivate,li2019practical} and every party is able to update the global parameters directly.

Google Keyboard~\cite{hard2018federated} is a case of centralized architecture. The server collects local model updates from users' devices and trains a global model, which is sent back to the users for inference, as shown in Figure \ref{fig:sgd_frame}. The scalability and stability are two important factors in the system design of the centralized FL. 

While the centralized design is widely used in existing studies, the decentralized design is preferred at some aspects since concentrating information on one server may bring potential risks or unfairness. \rev{However, the design of the decentralized communication architecture is challenging, which should take fairness and communication overhead into consideration. There are currently three different decentralized designs: P2P \cite{li2019practical,zhao2018inprivate}, graph \cite{marfoq2020throughput} or blockchain \cite{wang2021blockchain,zhao2020privacy}. In a P2P design, the parties are equally privileged and treated during federated learning. An example is SimFL \cite{li2019practical}, where each party trains a tree sequentially and sends the tree to all the other parties. The communication architecture can also be modeled as a graph with the additional constrains such as latency and computation time. \citet{marfoq2020throughput} propose an algorithm to find a throughput-optimal topology design. }Recently, blockchain~\cite{zheng2018blockchain} is a popular decentralized platform for consideration. It can be used to store the information of parties in federated learning and ensure the transparency of federated learning \cite{wang2021blockchain}. 

\subsection{Scale of Federation}
The FLSs can be categorized into two typical types by the scale of federation: cross-silo FLSs and cross-device FLSs \cite{kairouz2019advances}. The differences between them lie on the number of parties and the amount of data stored in each party.

In cross-silo FLS, the parties are organizations or data centers. There are usually a relatively small number of parties and each of them has a relatively large amount of data as well as computational power. For example, Amazon wants to recommend items for users by training the shopping data collected from hundreds of data centers around the world. Each data center possesses a huge amount of data as well as sufficient computational resources. \rev{Another example is that federated learning can be used among medical institutions. Different hospitals can use federated learning to train a CNN for chest radiography classification while keeping their chest X-ray images locally \cite{2021nature}. With federated learning, the accuracy of the model can be significantly improved.} One challenge that such FLS faces is how to efficiently distribute computation to data centers under the constraint of privacy models~\cite{zhou2019privacy}.

In cross-device FLS, on the contrary, the number of parties is relatively large and each party has a relatively small amount of data as well as computational power. The parties are usually mobile devices. Google Keyboard~\cite{yang2018applied} is an example of cross-device FLSs. The query suggestions of Google Keyboard can be improved with the help of FL. Due to the energy consumption concern, the devices cannot be asked to conduct complex training tasks. Under this occasion, the system should be powerful enough to manage a large number of parties and deal with possible issues such as the unstable connection between the device and the server. 

\subsection{Motivation of Federation}
\label{sec:motiv}
In real-world applications of FL, individual parties need the motivation to get involved in the FLS. The motivation can be regulations or incentives. FL inside a company or an organization is usually motivated by regulations (e.g., FL across different departments of a company). \rev{For example, the department which has the transaction records of users can help another department to predict user credit by federated learning.} In many cases, parties cannot be forced to provide their data by regulations. \rev{However, parties that choose to participate in federated learning can benefit from it, e.g., higher model accuracy. For example, hospitals can conduct federated learning to train a machine learning model for chest radiography classification \cite{2021nature} or COVID-19 detection \cite{qayyum2021collaborative}. Then, the hospitals can get a good model which has a higher accuracy than human experts and the model trained locally without federation.} 
Another example is Google Keyboard~\cite{yang2018applied}. While users have the choice to prevent Google Keyboard from utilizing their data, those who agree to upload input data may enjoy a higher accuracy of word prediction. Users may be willing to participate in federated learning for their convenience. 

A challenging problem is how to design a fair incentive mechanism, such that the party that contributes more can also benefit more from federated learning.
\REF{There have been some successful cases for incentive designs in blockchain~\cite{7163223,194906}. The parties inside the system can be collaborators as well as competitors.  \NBP{Other incentive designs like \cite{kang2019incentive2,kang2019incentive} are proposed to attract participants with high-quality data for FL.} We expect game theory models~\cite{sarikaya2019motivating,1210263,Naor:1999:PPA:336992.337028} and their equilibrium designs should be revisited under the FLSs. Even in the case of Google Keyboard, the users need to be motivated to participate this collaborative learning process. }

\section{Summary of Existing Studies}
\label{sec:exist_study}
In this section\footnote{Last updated on \today. We will periodically update this section to include the state-of-the-art and valuable FL studies. Please check out our latest version at this URL: https://arxiv.org/abs/1907.09693. Also, if you have any reference that you want to add into this survey, kindly drop Dr. Bingsheng He an email (hebs@comp.nus.edu.sg).}, we summarize and compare the existing studies on FLSs according to the aspects considered in Section~\ref{sec:taxonomy}. 

\subsection{Methodology}
To discover the existing studies on FL, we search keyword ``Federated Learning" in Google Scholar. Here we only consider the published studies in the computer science community.
%  and arXiv\footnote{\url{https://arxiv.org/}}

Since the scale of federation and the motivation of federation are problem dependent, we do not compare the existing studies by these two aspects. For ease of presentation, we use ``NN'', ``DT'' and ``LM'' to denote neural networks, decision trees and linear models, respectively. Moreover, we use ``CM'' and ``DP'' to denote cryptographic methods and differential privacy, respectively. Note that the algorithms (e.g., federated stochastic gradient descent) in some studies can be used to learn many machine learning models (e.g., logistic regression and neural networks). Thus, in the ``model implementation'' column, we present the models implemented in the experiments of corresponding papers. Moreover, in the ``main area'' column, we indicate the major area that the papers study on.

\subsection{Individual Studies}
\label{sec:ind_stu}

We summarize existing popular and state-of-the-art research work, as shown in Table~\ref{tbl:study_compare}. From Table~\ref{tbl:study_compare}, we have the following four key findings. 

First, most of the existing studies consider a horizontal data partitioning. We conjecture a part of the reason is that the experimental studies and benchmarks in horizontal data partitioning are relatively ready than vertical data partitioning. However, vertical FL is also common in real world, especially between different organizations. Vertical FL can enable more collaboration between diverse parties. Thus, more efforts should be paid to vertical FL to fill the gap.

Second, most studies consider exchanging the raw model parameters without any privacy guarantees. This may not be right if more powerful attacks on machine learning models are discovered in the future. Currently, the mainstream methods to provide privacy guarantees are differential privacy and cryptographic methods (e.g., secure multi-party computation and homomorphic encryption). Differential privacy may influence the final model quality a lot. Moreover, the cryptographic methods bring much computation and communication overhead and may be the bottleneck of FLSs. We look forward to a cheap way with reasonable privacy guarantees to satisfy the regulations.

Third, the centralized design is the mainstream of current implementations. A trusted server is needed in their settings. However, it may be hard to find a trusted server especially in the cross-silo setting. One naive approach to remove the central server is that the parties share the model parameters with all the other parties and each party also maintains the same global model locally. This method bring more communication and computation cost compared with the centralized setting. More studies should be done for practical FL with the decentralized architecture.

Last, the main research directions (also the main challenge) of FL are to improve the effectiveness, efficiency, and privacy, which are also three important metrics to evaluate an FLS. Meanwhile, there are many other research topics on FL such as fairness and incentive mechanisms. Since FL is related to many research areas, we believe that FL will attract more researchers and we can see more interesting studies in the near future.

\begin{table*}
\centering
\caption{Comparison among existing published studies. LM denotes Linear Models. DM denotes Decision Trees. NN denotes Neural Networks. CM denotes Cryptographic Methods. DP denotes Differential Privacy.}
\label{tbl:study_compare}
\resizebox{\textwidth}{!}{%
\begin{tabular}{|c|c|c|c|c|c|c|}
\hline
\begin{tabular}[c]{@{}c@{}}FL\\ Studies\end{tabular} &
\begin{tabular}[c]{@{}c@{}}main\\ area\end{tabular} &
\begin{tabular}[c]{@{}c@{}}data\\ partitioning\end{tabular} & \begin{tabular}[c]{@{}c@{}}model\\ implementation\end{tabular} & \begin{tabular}[c]{@{}c@{}}privacy\\ mechanism\end{tabular} & \begin{tabular}[c]{@{}c@{}}communication\\ architecture\end{tabular} &
\begin{tabular}[c]{@{}c@{}}remark\end{tabular}\\ \hline
FedAvg~\cite{mcmahan2016communication} & \multirow{27}{*}{\begin{tabular}[c]{@{}c@{}}Effective \\Algorithms\end{tabular}} & \multirow{12}{*}{horizontal} & NN & \multirow{17}{*}{\backslashbox{}{}} & \multirow{16}{*}{centralized} &\multirow{13}{*}{\begin{tabular}[c]{@{}c@{}}SGD-based \end{tabular}}\\\cline{1-1}\cline{4-4}
FedSVRG~\cite{konevcny2016federated2} & & &LM & & & \\ \cline{1-1}\cline{4-4}
FedProx~\cite{li2018federated} & & & LM, NN &  & & \\ \cline{1-1}\cline{4-4}
SCAFFOLD~\cite{karimireddy2020scaffold} & & & LM, NN & & & \\ \cline{1-1}\cline{4-4}
FedNova~\cite{wang2020tackling} & & & NN & & & \\ \cline{1-1}\cline{4-4}
Per-FedAvg \cite{perfedavg} & & & NN & & & \\ \cline{1-1}\cline{4-4}
pFedMe \cite{pFedMe} & & & LM, NN & & & \\ \cline{1-1}\cline{4-4}
IAPGD, AL2SGD+ \cite{apgd} & & & LM & & & \\ \cline{1-1}\cline{4-4}
IFCA \cite{ifca} & & & LM, NN & & & \\ \cline{1-1}\cline{4-4}
Agnostic FL~\cite{mohri2019agnostic} &  &  &LM, NN  & & &\\\cline{1-1}\cline{4-4}
FedRobust \cite{fedrobust} &  &  &NN  & & &\\\cline{1-1}\cline{4-4}
FedDF \cite{lin2020ensemble} &  &  &NN  & & &\\\cline{1-1}\cdashline{2-2}\cline{3-3}\cline{4-4}
FedBCD~\cite{liu2019communication} & &vertical & \multirow{4}{*}{NN} &  & & \\ \cline{1-1}\cline{3-3}\cdashline{4-6}\cline{7-7}
PNFM \cite{yurochkin2019bayesian} & & \multirow{2}{*}{horizontal} &  &  &&\multirow{3}{*}{\begin{tabular}[c]{@{}c@{}}NN-specialized\end{tabular}} \\\cline{1-1}
FedMA \cite{wang2020federated} & & &  & & & \\ \cline{1-1}\cline{3-3}
SplitNN \cite{wang2020federated} & &vertical & && & \\ \cline{1-1}\cdashline{2-2}\cline{3-3}\cline{4-7}
Tree-based FL~\cite{zhao2018inprivate} & &\multirow{4}{*}{horizontal} &\multirow{5}{*}{DT}  &DP & \multirow{2}{*}{decentralized} &\multirow{5}{*}{DT-specialized}\\ \cline{1-1}\cline{5-5}
SimFL~\cite{li2019practical} & & & & hashing & &\\\cline{1-1}\cdashline{2-4}\cline{5-6}
FedXGB~\cite{liu2019boosting} & &   & &\multirow{7}{*}{CM} &\multirow{16}{*}{centralized} & \\\cline{1-1}
FedForest \cite{liu2019federated} & & &  &  &  &\\\cline{1-1}\cline{3-3}
SecureBoost~\cite{cheng2019secureboost} & &vertical  &  & &  & \\\cline{1-1}\cdashline{2-2}\cline{3-4}\cdashline{5-6}\cline{7-7}
Ridge Regression FL~\cite{nikolaenko2013privacy} &  &\multirow{2}{*}{horizontal}  & \multirow{5}{*}{LM} & &  &\multirow{4}{*}{LM-specialized} \\\cline{1-1}
PPRR~\cite{chen2018privacy} & &  &  & &  & \\\cline{1-1}\cline{3-3}
Linear Regression FL~\cite{sanil2004privacy}  & & vertical & & & &\\\cline{1-1}\cline{3-3}
Logistic Regression FL~\cite{hardy2017private} &&\multirow{32}{*}{horizontal} & & & &\\\cline{1-1}\cline{5-5}\cline{7-7}\cdashline{2-4}\cdashline{6-6}
Federated MTL~\cite{smith2017federated} & &  &  &\multirow{9}{*}{\backslashbox{}{}} &  &multi-task learning \\\cline{1-1}\cline{4-4}\cline{7-7}\cdashline{2-3}\cdashline{5-6}
Federated Meta-Learning~\cite{chen2018federated} & &  &\multirow{3}{*}{NN}  &  & &\multirow{2}{*}{meta-learning} \\\cline{1-1}
Personalized FedAvg~\cite{jiang2019improving} & & &  &  &  & \\ \cline{1-1}\cline{7-7}\cdashline{2-6}
LFRL~\cite{liu2019lifelong} & &  &  & & &reinforcement learning\\\cline{1-1}\cline{4-4}\cline{7-7}
FBO~\cite{dai2020federated} &&&LM&&&Bayesian optimization \\ \cline{1-2} \cline{4-4}\cline{7-7}\cdashline{3-3}\cdashline{5-6}
Structure Updates~\cite{konevcny2016federated} &\multirow{19}{*}{\begin{tabular}[c]{@{}c@{}}Practicality\\Enhancement\end{tabular}} &  &\multirow{7}{*}{NN}  &  &  &\multirow{5}{*}{\begin{tabular}[c]{@{}c@{}}efficiency\\ improvement\end{tabular}} \\\cline{1-1}
Multi-Objective FL~\cite{zhu2019multi} &  &  &  &  & & \\\cline{1-1}
On-Device ML~\cite{jeong2018communication} & &  &  & & &\\\cline{1-1}
Sparse Ternary Compression~\cite{sattler2019robust} & &  &  & & &\\\cline{1-1}\cline{6-6}\cdashline{2-5}
DPASGD~\cite{marfoq2020throughput} & &  &  & &decentralized &\\\cline{1-1}\cline{5-7}\cdashline{2-4}

Client-Level DP FL~\cite{geyer2017differentially} & &  &  &\multirow{3}{*}{DP} & \multirow{24}{*}{centralized}  &\multirow{5}{*}{\begin{tabular}[c]{@{}c@{}}privacy\\guarantees\end{tabular}}\\\cline{1-1}
FL-LSTM~\cite{mcmahan2017learning} & &  &  & & &\\\cline{1-1}\cline{4-4}
Local DP FL~\cite{bhowmick2018protection} & & &LM, NN & & &\\\cline{1-1}\cline{4-5}
Secure Aggregation FL~\cite{bonawitz2017practical} & &  &NN &CM & & \\\cline{1-1}\cline{4-5}
Hybrid FL~\cite{truex2019hybrid} & &  &LM, DT, NN & CM, DP & &\\\cline{1-1}\cline{4-5}\cline{7-7}\cdashline{2-3}\cdashline{6-6}
Backdoor FL \cite{bagdasaryan2020backdoor,sun2019can,wang2020attack} & & &\multirow{4}{*}{NN}  & \multirow{13}{*}{\backslashbox{}{}} & &\multirow{6}{*}{robustness and attacks}\\\cline{1-1}
Adversarial Lens \cite{bhagoji2018analyzing} & & & & & &\\\cline{1-1}
Distributed Backdoor \cite{xie2019dba} & & & & & &\\\cline{1-1}
Image Reconstruction \cite{geiping2020inverting} & & & & & &  \\\cline{1-1}\cline{4-4}\cdashline{2-3}
RSA \cite{li2019rsa} & & &LM & & & \\ \cline{1-1}\cline{4-4}
Model Poison \cite{fang2020local} & & &LM, NN & & & \\ \cline{1-1}\cline{4-4}\cline{7-7}\cdashline{5-6}\cdashline{2-3}
$q$-FedAvg \cite{li2019fair} & & &LM, NN & & &fairness\\\cline{1-1}\cline{4-4}\cline{7-7}
BlockFL~\cite{kim2018device} & &  &\multirow{2}{*}{LM}  & & &\multirow{2}{*}{incentives}\\\cline{1-1}
Reputation FL \cite{kang2019incentive} & & & &  & & \\\cline{1-2}\cline{4-4}\cline{7-7}\cdashline{3-3}\cdashline{5-6}
FedCS~\cite{nishio2019client} &\multirow{9}{*}{Applications} &  &\multirow{2}{*}{NN}  & &  &\multirow{4}{*}{edge computing}\\\cline{1-1}
DRL-MEC~\cite{wang2019edge} & &  &  & & &\\\cline{1-1}\cline{4-4}
Resource-Constrained MEC~\cite{wang2019adaptive}&  & &LM, NN & & &\\\cline{1-1}\cline{4-4}
FedGKT~\cite{fedgkt}&  & &NN & & &\\\cline{1-1}\cline{4-4}\cline{7-7}\cdashline{2-3}\cdashline{5-6}
FedCF~\cite{ammad2019federated} & & &\multirow{2}{*}{LM} & & &collaborative filter\\ \cline{1-1}\cline{7-7}
FedMF~\cite{chai2019secure} & & & & & &matrix factorization\\ \cline{1-1}\cline{4-5}\cline{7-7}\cdashline{2-3}\cdashline{6-6}
FedRecSys \cite{tan2020federated} & & &LM, NN & CM
& &recommender system \\\cline{1-1}\cline{4-5}\cline{7-7}
FL Keyboard~\cite{hard2018federated} & & &NN &\multirow{5}{*}{\backslashbox{}{}} & & natural language processing \\ \cline{1-1}\cline{4-4}\cline{7-7}
Fraud detection \cite{zheng2020federated} &  & &NN & & &credit card transaction \\ \cline{1-4}\cline{6-7}\cdashline{5-5}
FedML~\cite{he2020fedml} &\multirow{10}{*}{Benchmarks} &\begin{tabular}[c]{@{}c@{}}horizontal\\ \&vertical\end{tabular} & LM, NN &  & \begin{tabular}[c]{@{}c@{}}centralized\\\&decentralized\end{tabular} &\multirow{4}{*}{general purpose benchmarks}\\\cline{1-1} \cline{3-3}\cline{4-4} \cline{6-6}
FedEval \cite{chai2020fedeval} & &\multirow{9}{*}{horizontal} & NN & &centralized & \\\cline{1-1} \cline{4-6} 
OARF \cite{hu2020oarf} & &  & NN &CM,DP & centralized & \\\cline{1-1} \cline{4-4} \cline{5-6}
Edge AIBench \cite{hao2019edge} & &  & \backslashbox{}{} & \multirow{7}{*}{\backslashbox{}{}} &\backslashbox{}{}& \\\cline{1-1}\cline{4-4}\cline{6-7}\cdashline{2-3}\cdashline{5-5}
PerfEval \cite{nilsson2018performance} & &  & \multirow{3}{*}{NN} &  & \multirow{3}{*}{centralized} & \multirow{4}{*}{targeted benchmarks} \\ \cline{1-1}
FedReID \cite{zhuang2020performance} & &  &  &  &  & \\ \cline{1-1}
semi-supervised benchmark \cite{zhang2020benchmarking} & &  & &  &  & \\ \cline{1-1}\cline{4-4} \cline{6-6}
non-IID benchmark \cite{liu2020evaluation} & &  &  \multirow{3}{*}{\backslashbox{}{}} &  &  \backslashbox{}{}  & \\ \cline{1-1} \cline{6-6}\cline{7-7}\cdashline{2-5}
LEAF~\cite{caldas2018leaf} & &  & \backslashbox{}{} & &centralized &\multirow{2}{*}{datasets}\\\cline{1-1} \cline{6-6}
Street Dataset \cite{luo2019real} & &  && & \backslashbox{}{} & \\ \hline
\end{tabular}%

}
\end{table*}

\subsubsection{Effectiveness Improvement}
\label{sec:eff_alg}
While some algorithms are based on SGD, the other algorithms are specially designed for one or several kinds of model architectures. Thus, we classify them into SGD-based algorithms and model specialized algorithms accordingly. 

\subsubsection*{SGD-Based}
If we consider the local data on a party as a single batch, SGD can be easily implemented in a federated setting by performing a single batch gradient calculation each round (i.e., FedSGD \cite{mcmahan2016communication}). However, such method may require a large number of communication rounds to converge. To reduce the number of communication rounds, FedAvg~\cite{mcmahan2016communication}, as introduced in Section 2.3.3 and Figure 1a of the main paper, is now a typical and practical FL framework based on SGD. In FedAvg, each party conducts multiple training rounds with SGD on its local model. Then, the weights of the global model are updated as the mean of weights of the local models. The global model is sent back to the parties to finish a global iteration. By averaging the weights, the local parties can take multiple steps of gradient descent on their local models, so that the number of communication rounds can be reduced compared with FedSGD.

\citet{konevcny2016federated2} propose federated SVRG (FSVRG). The major difference between federated SVRG and federated averaging is the way to update parameters of the local model and global model (i.e., step 2 and step 4). The formulas to update the model weights are based on stochastic variance reduced gradient (SVRG)~\cite{johnson2013accelerating} and distributed approximate newton algorithm (DANE) in federated SVRG. They compare their algorithm with the other baselines like CoCoA+~\cite{ma2017distributed} and simple distributed gradient descent. Their method can achieve better accuracy with the same communication rounds for the logistic regression model. There is no comparison between federated averaging and federated SVRG.

\QBB{A key challenge in federated learning is the heterogeneity of local data (i.e., non-IID data) \cite{niidbench}, which can degrade the performance of federated learning a lot \cite{li2018federated,karimireddy2020scaffold,li2019convergence}. Since the local models are updated towards their local optima, which are far from each other due to non-IID data, the averaged global model may also far from the global optima. To address the challenge,~\citet{li2018federated} propose FedProx. Since too many local updates may lead the averaged model far from the global optima, FedProx introduces an additional proximal term in the local objective to limit the amount of local changes. Instead of directly limiting the size of local updates, SCAFFOLD \cite{karimireddy2020scaffold} applies the variance reduction technique to correct the local updates. While FedProx and SCAFFOLD improve the local training stage of FedAvg, FedNova \cite{wang2020tackling} improves the aggregation stage of FedAvg. It takes the heterogeneous local updates of each party into consideration and normalizes the local models according to the local updates before averaging.}

\QBB{The above studies' objective is to minimize the loss on the whole training dataset under the non-IID data setting. Another solution is to design personalized federated learning algorithms, where the aim is that each party learns a personalized model which can perform well on its local data. Per-FedAvg \cite{perfedavg} applies the idea of model-agnostic meta-learning \cite{finn2017model} framework in FedAvg. pFedMe \cite{pFedMe} uses Moreau envelope to help decompose the personalized model optimization. \citet{apgd} establish the lower bound for the communication complexity and local oracle complexity of the personalized federated learning optimization. Moreover, they apply accelerated proximal gradient descent (APGD) and accelerated L2SGD+ \cite{hanzely2020federated}, which can achieve optimal complexity bound. IFCA \cite{ifca} assumes that the parties are partitioned into clusters by the local objectives. The idea is to alternatively minimize the loss functions while estimating the cluster identities.}

\QBB{Another research direction related to the non-IID data setting is to design robust federated learning against possible combinations of the local distributions. \citet{mohri2019agnostic} propose a new framework named agnostic FL. Instead of minimizing the loss with respect to the average distribution among the data distributions from local clients, they try to train a centralized model optimized for any possible target distribution formed by a mixture of the client distributions. FedRobust \cite{fedrobust} considers a structured affine distribution shifts. It proposes gradient descent ascent method to solve the distributed minimax optimization problem.}
% robust federated learning

\QBB{While the above studies consider the heterogeneity of data, the heterogeneity of the local models may also exist in federated learning. The parties can train models with different architectures. FedDF \cite{lin2020ensemble} utilizes knowledge distillation \cite{hinton2015distilling} to aggregate the local models. It assumes a public dataset exists on the server-side, which can be used to extract the knowledge of the local models and update the global model.}

There are few studies on SGD-based vertical federated learning. \cite{liu2019communication} propose the Federated Stochastic Block Coordinate Descent (FedBCD) for vertical FL. By applying coordinate descent, each party updates its local parameter for multiple rounds before communicating the intermediate results. They also provide convergence analysis for FedBCD. \citet{hu2019fdml} propose FDML for vertical FL assuming all parties have the labels. Instead of exchanging the intermediate results, it aggregates the local prediction from each of the participated party.

\subsubsection*{Neural Networks}
Although neural networks can be trained using the SGD optimizer, we can potentially increase the model utility if the model architecture can also be exploited.~\citet{yurochkin2019bayesian} develop probabilistic federated neural matching (PFNM) for multilayer perceptrons by applying Bayesian nonparametric machinery~\cite{gershman2012tutorial}. They use an Beta-Bernoulli process informed matching procedure to combine the local models into a federated global model. The experiments show that their approach can outperform FedAvg on both IID and non-IID data partitioning.

\citet{wang2020federated} show how to apply PFNM to CNNs (convolutional neural networks) and LSTMs (long short-term memory networks). Moreover, they propose Federated Matched Averaging (FedMA) with a layer-wise matching scheme by exploting the model architecture. Specifically, they use matched averaging to update a layer of the global model each time, which also reduces the communication size. The experiments show that FedMA performs better than FedAvg and FedProx \cite{li2018federated} on CNNs and LSTMs.

\LX{Another study for vertical federated learning on neural networks is split learning \cite{vepakomma2018split}. \citet{vepakomma2018split} propose a novel paradigm named SplitNN, where a neural network is divided into two parts. Each participated party just need to train a few layers of the network, then the output at the cut layer are transmitted to the party who has label and completes the rest of the training.}

\subsubsection*{Trees}
Besides neural networks, decision trees are also widely used in the academic and industry~\cite{chen2016xgboost,Ke2017LightGBMAH,feng2018multi,li2019privacy}. Compared with NNs, the training and inference of trees are highly efficient. However, the tree parameters cannot be directly optimized by SGD, which means that SGD-based FL frameworks are not applicable to learn trees. We need specialized frameworks for trees. Among the tree models, the Gradient Boosting Decision Tree (GBDT) model~\cite{chen2016xgboost} is quite popular. There are several studies on federated GBDT. 

There are some studies on horizontal federated GBDTs. \citet{zhao2018inprivate} propose the first FLS for GBDTs. In their framework, each decision tree is trained locally without the communications between parties. The trees trained in a party are sent to the next party to continuous train a number of trees. Differential privacy is used to protect the decision trees.
\citet{li2019practical} exploit similarity information in the building of federated GBDTs by using locality-sensitive hashing~\cite{datar2004locality}. They utilize the data distribution of local parties by aggregating gradients of similar instances. Within a weaker privacy model compared with secure multi-party computation, their approach is effective and efficient.
\citet{liu2019boosting} propose a federated extreme boosting learning framework for mobile crowdsensing. They adopted secret sharing to achieve privacy-preserving learning of GBDTs.

\citet{liu2019federated} propose Federated Forest, which enables training random forests in the vertical FL setting. In the building of each node, the party with the corresponding split feature is responsible for splitting the samples and sharing the results. They encrypt the communicated data to protect privacy. Their approach is as accurate as the non-federated version.

\citet{cheng2019secureboost} propose SecureBoost, a framework for GBDTs in the vertical FL setting. In their assumption, only one party has the label information. They used the entity alignment technique to get the common data and then build the decision trees. Additively homomorphic encryption is used to protect the gradients.

\subsubsection*{Linear/Logistic Regression}
Linear/logistic regression can be achieved using SGD. Here we show the studies that are not SGD-based and specially designed for linear/logistic regression.

In the horizontal FL setting, \citet{nikolaenko2013privacy} propose a system for privacy-preserving ridge regression. Their approaches combine both homomorphic encryption and Yao's garbled circuit to achieve privacy requirements. An extra evaluator is needed to run the algorithm.
% linear regression, framework
\citet{chen2018privacy} propose a system for privacy-preserving ridge regression. Their approaches combine both secure summation and homomorphic encryption to achieve privacy requirements. They provided a complete communication and computation overhead comparison among their approach and the previous state-of-the-art approaches. 

In the vertical FL setting, \citet{sanil2004privacy} present a secure regression model. They focus on the linear regression model and secret sharing is applied to ensure privacy in their solution.
% linear regression, secure summation, vertical
\citet{hardy2017private} present a solution for two-party vertical federated logistic regression. They apply entity resolution and additively homomorphic encryption.

\subsubsection*{Others}
There are many studies that combine FL with other machine learning techniques such as multi-task learning~\cite{ruder2017overview}, meta-learning~\cite{finn2017model}, reinforcement learning~\cite{mnih2015human}, and transfer learning~\cite{pan2010survey}.

\citet{smith2017federated} combine FL with multi-task learning~\cite{caruana1997multitask,zhang2017survey}. Their method considers the issues of high communication cost, stragglers, and fault tolerance for MTL in the federated environment. 
% federated multi-task learning, algorithm: multi-task learning
\citet{corinzia2019variational} propose a federated MTL method with non-convex models. They treated the central server and the local parties as a Bayesian network and the inference is performed using variational methods.
% federated multi-task learning

\citet{chen2018federated} adopt meta-learning in the learning process of FedAvg. Instead of training the local NNs and exchanging the model parameters, the parties adopt the Model-Agnostic Meta-Learning (MAML)~\cite{finn2017model} algorithm in the local training and exchange the gradients of MAML. 
% federated meta-learning
\citet{jiang2019improving} interpret FedAvg in the light of existing MAML algorithms. Furthermore, they apply Reptile algorithm~\cite{nichol2018reptile} to fine-tune the global model trained by FedAvg. Their experiments show that the meta-learning algorithm can improve the effectiveness of the global model.

\citet{liu2019lifelong} propose a lifelong federated reinforcement learning framework. Adopting transfer learning techniques, a global model is trained to effectively remember what the robots have learned in reinforcement learning.
% federated reinforcement learning, framework

\citet{dai2020federated} considers Bayesian optimization in FL. They propose federated Thompson sampling to address the communication efficiency and heterogeneity of the clients. Their approach can potentially be used in the parameter search in federated learning.

Another issue in FL is the package loss or party disconnection during FL process, which usually happens on mobile devices. When the number of failed messages is small, the server can simply ignore them as they have a small weight on the updating of the global model. If the party failure is significant, the server can restart from the results of the previous round \cite{bonawitz2019towards}. We look forward to more novel solutions to deal with the disconnection issue for effectiveness improvement.

\subsubsection*{\tb{Summary}} We summarize the above studies as follows.
\begin{itemize}
    \item As the SGD-based framework has been widely studied and used, more studies focus on model specialized FL recently. We expect to achieve better model accuracy by using model specialized methods. Moreover, we encourage researchers to study on federated decision trees models. The tree models have a small model size and are easy to train compared with neural networks, which can result in a low communication and computation overhead in FL.
    
    \item The study on FL is still on a early stage. Few studies have been done on appling FL to train the state-of-the-art neural networks such as ResNeXt \cite{mahajan2018exploring} and EfficientNet \cite{tan2019efficientnet}. How to design an effective and practical algorithm to train a complex machine learning model is still a challenging and on-going research direction.
    
    \item While most studies focus on horizontal FL, there is still no well developed algorithm for vertical FL. However, the vertical federated setting is common in real world applications where multiple organizations are involved. We look forward to more studies on this promising area.
\end{itemize}

\subsubsection{Communication Efficiency}
\label{sec:comm}

While the computation of FL can be accelerated using modern hardware and techniques~\cite{lopes2011gpumlib,li2015heterospark,li2019adaptive} in high performance computing community~\cite{wen2018efficient,wen2019exploiting}, the FL studies mainly work on reducing the communication size during the FL process.

\citet{konevcny2016federated} propose two ways, structured updates and sketched updates, to reduce the communication costs in federated averaging. The first approach restricts the structure of local updates and transforms it to the multiplication of two smaller matrices. Only one small matrix is sent during the learning process. The second approach uses a lossy compression method to compress the updates. Their method can reduce the communication cost by two orders of magnitude with a slight degradation in convergence speed.
% FA, compression, efficiency improvement
\citet{zhu2019multi} design a multi-objective evolutionary algorithm to minimize the communication costs and global model test errors simultaneously. Considering the minimization of the communication cost and the maximization of the global learning accuracy as two objectives, they formulated FL as a bi-objective optimization problem and solve it by the multi-objective evolutionary algorithm.
% FA, efficiency improvement, two objectives
\citet{jeong2018communication} propose a FL framework for devices with non-IID local data. They design federated distillation, whose communication size depends on the output dimension but not on the model size. Also, they propose a data augmentation scheme using a generative adversarial network (GAN) to make the training dataset become IID. \NBP{Many other studies also design specialize approach for non-IID data~\cite{zhao2018federated,li2019convergence,liu2019edge,yoshida2019hybrid}}.
% Efficiecny improvement, federated distillation, co-distillation, gradient descent
\citet{sattler2019robust} propose a new compression framework named sparse ternary compression (STC). Specifically, STC compresses the communication using sparsification, ternarization, error accumulation, and optimal Golomb encoding. Their method is robust to non-IID data and large numbers of parties. 

\QBB{Beside the communication size, the communication architecture can also be improved to increase the training efficiency. \citet{marfoq2020throughput} consider the topology design for cross-silo federated learning. They propose an approach to find a throughput-optimal topology, which can significantly reduce the training time.}
% SGD, efficiency improvement, compression

\subsubsection{Privacy, Robustness and Attacks}
\label{sec:priv}

Although the original data is not exchanged in FL, the model parameters can also leak sensitive information about the training data \cite{shokri2017membership,nasr2019comprehensive,wang2019beyond}. Thus, it is important to provide privacy guarantees for the exchanged local updates.

Differential privacy is a popular method to provide privacy guarantees. \citet{geyer2017differentially} apply differential privacy in federated averaging from a client-level perspective. They use the Gaussian mechanism to distort the sum of updates of gradients to protect a whole client's dataset instead of a single data point.
% FA, privacy enhancement
\citet{mcmahan2017learning} deploy federated averaging in the training of LSTM. They also use client-level differential privacy to protect the parameters.
% DP-FedAvg, privacy enhancement
\citet{bhowmick2018protection} apply local differential privacy to protect the parameters in FL. To increase the model quality, they consider a practical threat model that wishes to decode individuals' data but has little prior information on them. Within this assumption, they can better utilize the privacy budget.
% privacy enhancement, sgd

\citet{bonawitz2017practical} apply secure multi-party computation to protect the local parameters on the basis of federated averaging. Specifically, they present a secure aggregation protocol to securely compute the sum of vectors based on secret sharing \cite{shamir1979share}. They also discuss how to combine differential privacy with secure aggregation. 

\citet{truex2019hybrid} combine both secure multiparty computation and differential privacy for privacy-preserving FL. They use differential privacy to inject noises to the local updates. Then the noisy updates will be encrypted using the Paillier cryptosystem~\cite{paillier1999public} before sent to the central server.

For the attacks on FL, one kind of popular attack is backdoor attack, which aims to achieve a bad global model by exchanging malicious local updates.

\citet{bagdasaryan2020backdoor} conduct model poisoning attack on FL. The malicious parties commit the attack models to the server so that the global model may overfit with the poisoned data. The secure multi-party computation cannot prevent such attack since it aims to protect the confidentiality of the model parameters.
\citet{bhagoji2018analyzing} also study the model poisoning attack on FL. Since the averaging step will reduce the effect of the malicious model, it adopts an explicit boosting way to increase the committed weight update.
\NB{\citet{sun2019can} conduct experiments to evaluate backdoor attacks and defenses for federated learning on federated EMNIST dataset to see what factors can affect the performance of adversary. They find that in the absence of defenses, the performance of the attack largely depends on the fraction of adversaries presented and the "complexity" of the targeted task. The more backdoor tasks we have, the harder it is to backdoor a ﬁxed-capacity model while maintaining its performance on the main task.} \LX{\citet{wang2020attack} discuss the backdoor attack from a theoretical view and prove that it is feasible in FL. They also propose a new class of backdoor attacks named edge-case backdoors, which are resistant to the current defending methods.} \citet{xie2019dba} propose a distributed backdoor attack on FL. They decompose the global trigger pattern into local patterns. Each adversarial party only employs one local pattern. The experiments show that their distributed backdoor attack outperforms the central backdoor attack.

\NB{Another kind of attack is the \textit{Byzantine}-attacks, where adversaries fully control some authenticated devices and behave arbitrarily to disrupt the network. There have been some existing robust aggregation rules in distributed learning such as $Krum$ \cite{blanchard2017machine} and Bulyan \cite{mhamdi2018hidden}. These rules can be directly applied in federated learning. However, since each party conduct multiple local update steps in federated learning, it is interesting to investigate the Byzantine attacks and defenses in federated learning. \citet{li2019rsa} propose RSA, a Byzantine-robust stochastic aggregation method for federated learning on non-IID data setting. \citet{fang2020local} propose model poison attacks for byzantine-robust federated learning approaches. The goal of their approach is to modify the local models such that the global model deviates the most towards the inverse of the correct update direction. }

\LY{Another line of study about FL attack are the inference attacks. There are existing studies for the inferences attack \cite{fredrikson2015model,shokri2017membership,nasr2019comprehensive} on the machine learning model trained in a centralized setting. For the federated setting, \citet{geiping2020inverting} show that it is possible to reconstruct the training images from the knowledge of the exchanged gradients.}

\subsubsection{Fairness and Incentive Mechanisms}
\label{sec:fair}
% Besides efficiency and privacy, there are also other aspects of FL can be improved such as fairness, energy saving, and incentive mechanisms.
By taking fairness into consideration based on FedAvg,~\citet{li2019fair} propose $q$-FedAvg. Specifically, they define the fairness according to the variance of the performance of the model on the parties. If such variance is smaller, then the model is more fair. Thus, they design a new objective inspired by $\alpha$-fairness~\cite{altman2008generalized}. Based on federated averaging, they propose $q$-FedAvg to solve their new objective. The major difference between $q$-FedAvg with FedAvg is in the formulas to update model parameters.\del{ The experimental results on a linear SVM and an LSTM showed that the performance variance is much smaller than the vanilla FedAvg. }

\citet{kim2018device} combine blockchain architecture with FL. On the basis of federated averaging, they use a blockchain network to exchange the devices' local model updates, which is more stable than a central server and can provide the rewards for the devices. \citet{kang2019incentive} designed a reputation-based worker selection scheme for reliable FL by using a multi-weight subjective logic model. They also leverage the blockchain to achieve secure reputation management for workers with non-repudiation and tamper-resistance properties in a decentralized manner.

\subsubsection*{\tb{Summary}} According to the review above, we summarize the studies in Section \ref{sec:comm} to Section \ref{sec:fair} as follows.
\begin{itemize}
    \item Besides effectiveness, efficiency and privacy are the other two important factors of an FLS. Compared with these three areas, there are fewer studies on fairness and incentive mechanisms. We look forward to more studies on fairness and incentive mechanisms, which can encourage the usage of FL in the real world.
    \item For the efficiency improvement of FLSs, the communication overhead is still the main challenge. Most studies \cite{konevcny2016federated,jeong2018communication,sattler2019robust} try to reduce the communication size of each iteration. How to reasonably set the number of communication rounds is also promising \cite{zhu2019multi}. The trade-off between the computation and communication still needs to be further investigated.
    \item For the privacy guarantees, differential privacy and secure multi-party computation are two popular techniques. However, differential privacy may impact the model quality significantly and secure multi-party computation may be very time-consuming. It is still challenging to design a practical FLS with strong privacy guarantees. Also, the effective robust algorithms against poisoning attacks are not widely adopted yet.
\end{itemize}

\subsubsection{Applications}
One related area with FL is edge computing \cite{niknam2019federated,yu2018federated,qian2019privacy1,duan2019astraea,zhao2019mobile}, where the parties are edge devices. Many studies try to integrate FL with the mobile edge systems. FL also shows promising results in recommender system \cite{ammad2019federated,chai2019secure,zhou2019fedrec}, natural language processing \cite{hard2018federated} \LX{and transaction fraud detection \cite{zheng2020federated}.}

\subsubsection*{Edge Computing}
\citet{nishio2019client} implement federated averaging in practical mobile edge computing (MEC) frameworks. They use an operator of MEC framworks to manage the resources of heterogeneous clients. \citet{wang2019edge} adopt both distributed deep reinforcement learning (DRL) and federatd learning in mobile edge computing system. The usage of DRL and FL can effectively optimize the mobile edge computing, caching, and communication. \citet{wang2019adaptive} perform FL on resource-constrained MEC systems. They address the problem of how to efficiently utilize the limited computation and communication resources at the edge. Using federated averaging, they implement many machine learning algorithms including linear regression, SVM, and CNN. \citet{fedgkt} also consider the limited computing resources in the edge devices. They propose FedGKT, where each device only trains a small part of a whole ResNet to reduce the computation overhead. 

\subsubsection*{Recommender System}\hfill\\
\citet{ammad2019federated} formulate the first federated collaborative filter method. Based on a stochastic gradient approach, the item-factor matrix is trained in a global server by aggregating the local updates. They empirically show that the federated method has almost no accuracy loss compared with the centralized method. \citet{chai2019secure} design a federated matrix factorization framework. They use federated SGD to learn the matrices. Moreover, they adopt homomorphic encryption to protect the communicated gradients. \LX{\citet{tan2020federated} build a federated recommender system (FedRecSys) based on FATE. FedRecSys has implemented popular recommendation algorithms with SMC protocols. The algorithms include matrix factorization, singular value decomposition, factorization machine, and deep learning.}

% \subsubsection*{Computer Vision}
% photolabel

\subsubsection*{Natural Language Processing}
\citet{hard2018federated} apply FL in mobile keyboard next-word prediction. They adopt the federated averaging method to learn a variant of LSTM called Coupled Input and Forget Gate (CIFG) \cite{greff2016lstm}. The FL method can achieve better precision recall than the server-based training with log data.
% google keyboard

\LX{
\subsubsection*{Transaction Fraud Detection}
\citet{zheng2020federated} introduce FL into the field of fraud detection on credit card transaction. They design a novel meta-learning based federated learning framework, named deep K-tuplet network, which not only guarantees data privacy but also achieves a significantly higher performance compared with the existing approaches.}

\subsubsection*{\tb{Summary}} According to the above studies, we have the following summaries.

\begin{itemize}
    \item Edge computing naturally fits the cross-device federated setting. A nontrivial issue of applying FL to edge computing is how to effectively utilize and manage the edge resources. The usage of FL can bring benefits to users, especially for improving mobile device services.
    \item FL can solve many traditional machine learning tasks such as image classification and work prediction. Due to the regulations and ``data islands'', the federated setting may be a common setting in the next years. With the fast development of FL, we believe that there will be more applications in computer vision, natural language processing, and healthcare.
\end{itemize}

\subsubsection{Benchmark}

Benchmark is important for directing the development of FLSs. Multiple benchmark-related works have been conducted recently, and several benchmark frameworks are available online. We categorize them into three types: 1) \emph{General purpose benchmark systems} aim at comprehensively evaluate FLSs and give a detailed characterization of different aspects of FLSs; 2) \emph{Targeted benchmarks} aim at one or more aspects that concentrated in a small domain and tries to optimize the performance of the system in that domain; 3) \emph{Dataset benchmarks} aim at providing dedicated datasets for federated learning.

\subsubsection*{General Purpose Benchmark Systems}
FedML~\cite{he2020fedml} is a research library that provides both frameworks for federated learning and benchmark functionalities. As a benchmark, it provides comprehensive baseline implementations for multiple ML models and FL algorithms, including FedAvg, FedNAS, Vertical FL, and split learning. Moreover, it supports three computing paradigms, namely distributed training, mobile on-device training, and standalone simulation. Although some of its experiment results are currently still at a preliminary stage, it is one of the most comprehensive benchmark frameworks concerning its functionalities.

FedEval~\cite{chai2020fedeval} is another evaluation model for federated learning. It features the ``ACTPR'' model, i.e., using accuracy, communication, time consumption, privacy and robustness as its evaluation targets. It utilizes Docker containers to provide an isolated evaluation environment to work around the hardware resource limitation problem, and simulated up to 100 clients in the implementation. Currently, two horizontal algorithms are supported: FedSGD and FedAvg, and the models including MLP and LeNet are tested.

OARF~\cite{hu2020oarf} provides a set of utilities and reference implementations for FL benchmarks. It features the measurement of different components in FLSs, including FL algorithms, encryption mechanisms, privacy mechanisms, and communication methods. In addition, it also features realistic partitioning of datasets, which utilizes public datasets collected from different sources to reflect real-world data distributions. Both horizontal vertical algorithms are tested.

Edge AIBench~\cite{hao2019edge} provides a testbed for federated learning applications, and models four application scenarios as reference implementations: ICU patients monitor, surveillance camera, smart home, and autonomous vehicles. The implementation is open sourced, but no experiment result has been reported currently.

\subsubsection*{Targeted Benchmarks}
\citet{nilsson2018performance} propose a method utilizing correlated t-test to compare between different types of federated learning algorithms while bypassing the influence of data distributions. Three FL algorithms, FedAvg, FedSVRG~\cite{konevcny2016federated} and CO-OP~\cite{wang2017co} are compared in both IID and non-IID setup in their work, and the result shows that FedAvg achieves the highest accuracy among the three algorithms regardless of how data is partitioned.

\citet{zhuang2020performance} utilize benchmark analysis to improve the performance of federated person re-identification. The benchmark part uses 9 different datasets to simulate real-world situations and uses federated partial averaging, an algorithm that allows the aggregation of partially different models, as the reference implementations. 

\citet{zhang2020benchmarking} present a benchmark targeted at semi-supervised federated learning setting, where users only have unlabelled data, and the server only has a small amount of labelled data, and explore the relation between final model accuracy and multiple metrics, including the distribution of the data, the algorithm and communication settings, and the number of clients. Utilizing the experiment results, their semi-supervised learning improved method achieves better generalization performance.

\citet{liu2020evaluation} focus on the non-IID problem, where datasets are distributed unevenly across the participating parties. Their work explores methods for quantitatively describing the skewness of the data distribution, and propose several non-IID dataset generation approaches. 

\subsubsection*{Datasets}
LEAF~\cite{caldas2018leaf} is one of the earliest dataset proposals for federated learning. It contains six datasets covering different domains, including image classification, sentiment analysis, and next-character prediction. A set of utilities is provided to divide datasets into different parties in an IID or non-IID way. For each dataset, a reference implementation is also provided to demonstrate the usage of that dataset in the training process.

\citet{luo2019real} present real-world image datasets which are collected from 26 different street cameras. Images in that dataset contain objects of 7 different categories and are suitable for the object detection task. Implementations with federated averaging running YOLOv3 model and Faster R-CNN model are provided as references.

\subsubsection*{\tb{Summary}}
Summarizing the studies above, we have the following discoveries 
\begin{itemize}
\item Benchmarks serve an important role in the development of federated learning. Through different types of benchmarks, we can quantitatively characterize the different components and aspects of federated learning. Benchmarks regarding the security and privacy issues in federated learning are still at an early stage and require further development.
\item Currently no comprehensive enough benchmark system has been implemented to cover all the algorithms or application types in FLSs. Even the most comprehensive benchmark systems lack supports for certain algorithms and evaluation metrics for each level of the system. Further development of comprehensive benchmark systems requires the support of extensive FL frameworks. 
\item Most benchmark researches are using datasets which are split from a single dataset, and there is no consensus on what type of splitting method should be used. Similarly, regarding the non-IID problem, there is no consensus on the metric of non-IID-ness. Using realistic partitioning method, as proposed in FedML~\cite{he2020fedml} and OARF~\cite{hu2020oarf} may mitigate this issue, but for federated learning at a large-scale, realistic partitioning is not suitable due to the difficulty of collecting data from different sources.
\end{itemize}

\subsection{Open Source Systems}
In this section, we introduce five open source FLSs: Federated AI Technology Enabler (FATE)\footnote{\url{https://github.com/FederatedAI/FATE}}, Google TensorFlow Federated (TFF)\footnote{\url{https://github.com/tensorflow/federated}}, OpenMined PySyft\footnote{\url{https://github.com/OpenMined/PySyft}}, Baidu PaddleFL\footnote{\url{https://github.com/PaddlePaddle/PaddleFL}}, and
FedML\footnote{\url{https://github.com/FedML-AI/FedML}}.

\subsubsection{FATE}
\begin{figure}
\begin{center}
\includegraphics[width=0.99\columnwidth]{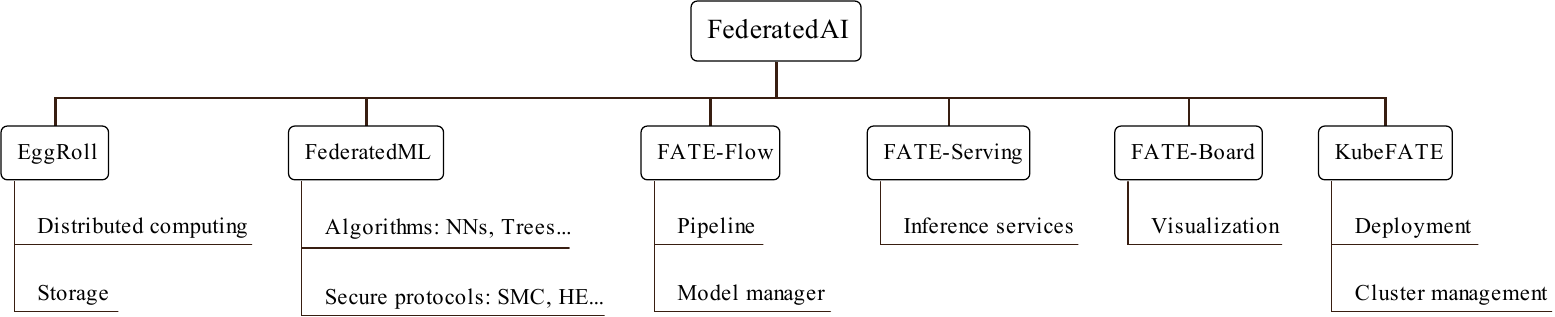}
\caption{The FATE system structure}
\label{fig:fate}
\end{center}
\end{figure}

\rev{FATE is an industrial level FL framework developed by WeBank, which aims to provide FL services between different organizations. FATE is based on Python and can be installed on Linux or Mac. It has attracted about 3.2k stars and 900 forks on GitHub.} The overall structure of FATE is shown in Figure \ref{fig:fate}. It has six major modules: EggRoll, FederatedML, FATE-Flow, FATE-Serving, FATE-Board, and KubeFATE. EggRoll manages the distributed computing and storage. It provides computing and storage AIPs for the other modules. FederatedML includes the federated algorithms and secure protocols. Currently, it supports training many kinds of machine learning models under both horizontal and vertical federated setting, including NNs, GBDTs, and logistic regression. \rev{FATE assumes that the parties are honest-but-curious. Thus, it uses secure multi-party computation and homomorphic encryption to protect the communicated messages. However, it does not support differential privacy to protect the final model.} FATE-Flow is a platform for the users to define their pipeline of the FL process. The pipeline can include the data preprocessing, federated training, federated evaluation, model management, and model publishing. FATE-Serving provides inference services for the users. It supports loading the FL models and conducting online inference on them. FATE-Board is a visualization tool for FATE. It provides a visual way to track the job execution and model performance. Last, KubeFATE helps deploy FATE on clusters by using Docker or Kubernetes. It provides customized deployment and cluster management services. In general, FATE is a powerful and easy-to-use FLS. Users can simply set the parameters to run a FL algorithm. Moreover, FATE provides detailed documents on its deployment and usage. However, since FATE provides algorithm-level interfaces, practitioners have to modify the source code of FATE to implement their own federated algorithms. This is not easy for non-expert users.

\subsubsection{TFF}
\begin{figure}
\begin{center}
\includegraphics[width=0.8\columnwidth]{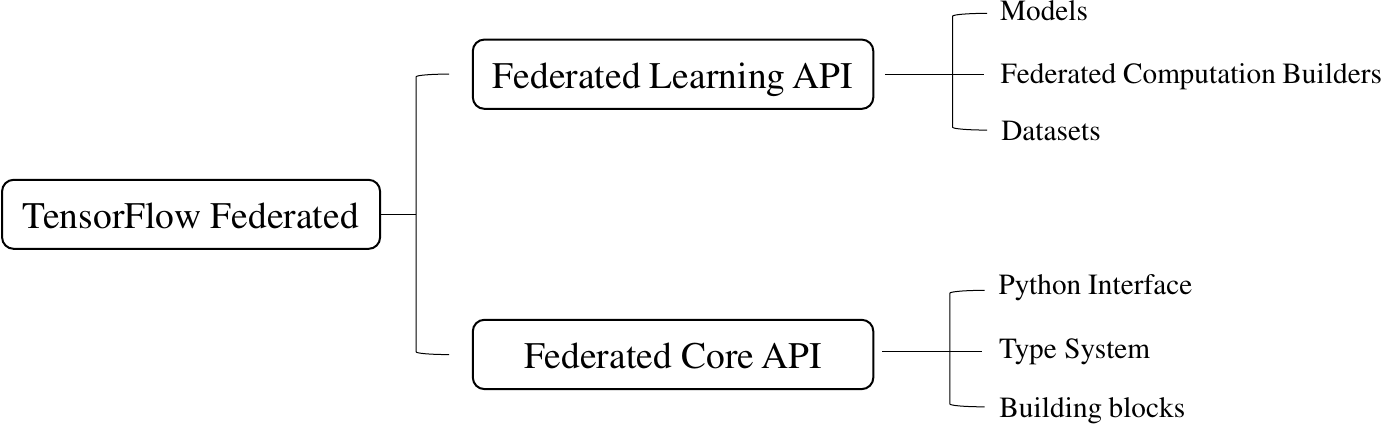}
\caption{The TFF system structure}
\label{fig:tff}
\end{center}
\end{figure}

\rev{TFF, developed by Google, provides the building blocks for FL based on TensorFlow. It has attracted about 1.5k stars and 380 forks on GitHub. TFF provides a Python package which can be easily installed and imported.} As shown in Figure~\ref{fig:tff}, it provides two APIs of different layers: FL API and Federated Core (FC) API. FL API offers high-level interfaces. It includes three key parts, which are models, federated computation builders, and datasets. FL API allows users to define the models or simply load the Keras~\cite{gulli2017deep} model. The federated computation builders include the typical federated averaging algorithm. Also, FL API provides simulated federated datasets and functions to access and enumerate the local datasets for FL. Besides high-level interfaces, FC API also includes lower-level interfaces as the foundation of the FL process. Developers can implement their functions and interfaces inside the federated core.
Finally, FC provides the building blocks for FL. It support multiple federated operators such as federated sum, federated reduce, and federated broadcast. Developers can define their own operators to implement the FL algorithm. Overall, TFF is a lightweight system for developers to design and implement new FL algorithms. Currently, \rev{TFF does not consider consider any adversaries during FL training. It does not provide privacy mechanisms.} TFF can only deploy on a single machine now, where the federated setting is implemented by simulation.

\subsubsection{PySyft}
\rev{PySyft, first proposed by \citet{ryffel2018generic} and developed by OpenMined, is a python library that provides interfaces for developers to implement their training algorithm. It has attracted about 7.3k stars and 1.7k forks on GitHub. While TFF is based on TensorFlow, PySyft can work well with both PyTorch and TensorFlow. PySyft provides multiple optional privacy mechanisms including secure multi-party computation and differential privacy. Thus, it can support running on honest-but-curious parties.} Moreover, it can be deployed on a single machine or multiple machines, where the communication between different clients is through the websocket API~\cite{skvorc2014performance}. However, while PySyft provides a set of tutorials, there is no detailed document on its interfaces and system architecture.

\subsubsection{PaddleFL}
\begin{figure}
\begin{center}
\includegraphics[width=0.8\columnwidth]{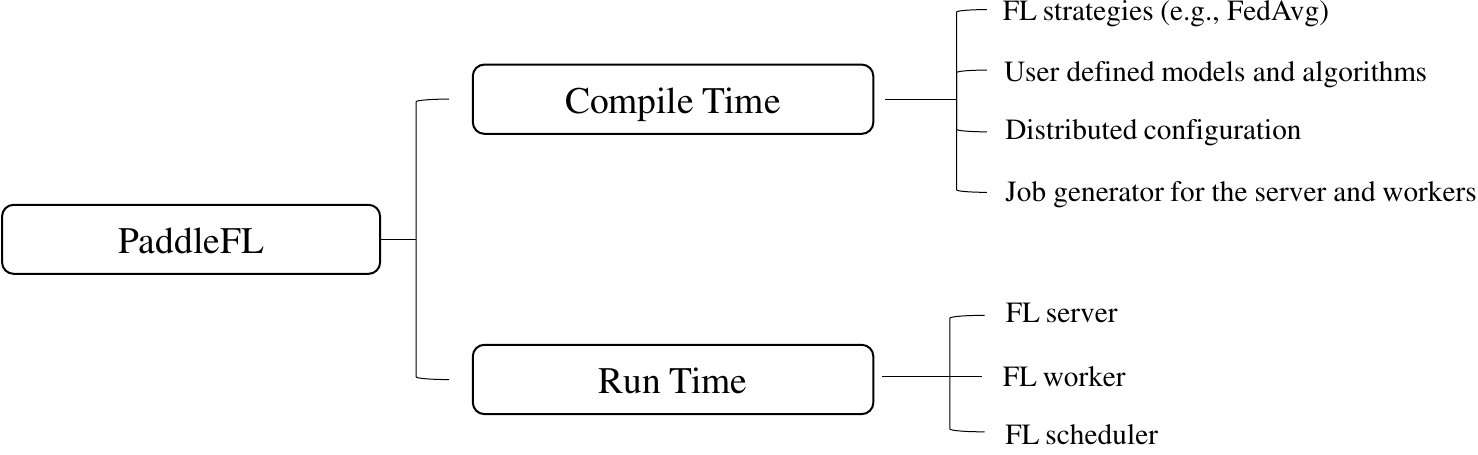}
\caption{The PaddleFL system structure}
\label{fig:paddlefl}
\end{center}
\end{figure}

\rev{PaddleFL is a FLS based on PaddlePaddle\footnote{\url{https://github.com/PaddlePaddle/Paddle}}, which is a deep learning platform developed by Baidu. It is implemented on C++ and Python. It has attracted about 260 stars and 60 forks on GitHub. Like PySyft, PaddleFL supports both differential privacy and secure multi-party computation and can work on honest-but-curious parties.} The system structure of PaddleFL is shown in Figure \ref{fig:paddlefl}. In the compile time, there are four components including FL strategies, user defined models and algorithms, distributed training configuration, and FL job generator. The FL strategies include the horizontal FL algorithms such as FedAvg. Vertical FL algorithms will be integrated in the future. Besides the provided FL strategies, users can also define their own models and training algorithms. The distributed training configuration defines the training node information in the distributed setting. FL job generator generates the jobs for federated server and workers. In the run time, there are three components including FL server, FL worker, and FL scheduler. The server and worker are the manager and parties in FL, respectively. The scheduler selects the workers that participate in the training in each round. Currently, the development of PaddleFL is still in a early stage and the documents and examples are not clear enough. 

\subsubsection{FedML}
\begin{figure}
\begin{center}
\includegraphics[width=0.8\columnwidth]{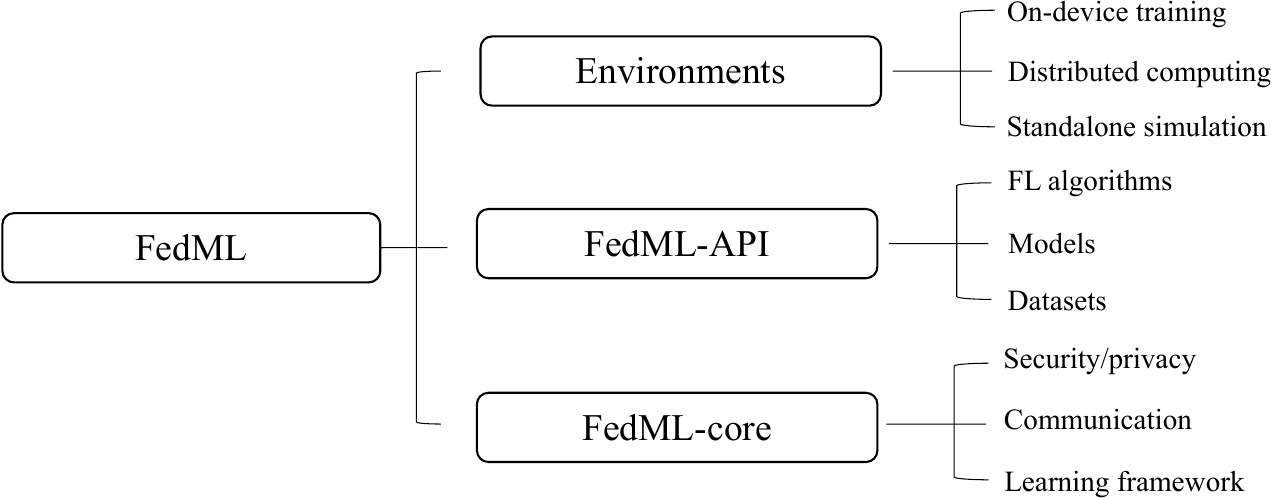}
\caption{The FedML system structure}
\label{fig:fedml}
\end{center}
\end{figure}

\SXH{\rev{FedML provides both a framework for federated learning and a platform for FL benchmark. It is developed by a team from University of Southern California \cite{he2020fedml} based on PyTorch. FedML has attracted about 660 stars and 180 forks on GitHub.} As an FL framework, It's core structure is divided into two levels, as shown in Figure~\ref{fig:fedml}. In the low-level FedML-core, training engine and distributed communication infrastructures are implemented. 
The high-level FedML-API is built on top of them and provides training models, datasets, and FL algorithms. Reference application/benchmark implementations are further built on top of the FedML-API. \rev{While most algorithms implemented on FedML does not consider any adversaries, it supports applying differential privacy when aggregating the messages from the parties.} FedML supports three computing paradigms, namely standalone simulation, distributed computing and on-device training, which provides a simulation environment for a broad spectrum of hardware requirements. Reference implementations for all supported FL algorithms are provided. Although there are still gaps between some of the experiment results and the optimal results, they  provide useful information for further development.}

\subsubsection{Others}
There are other closed source federated learning systems. NVIDIA Clara \footnote{\url{https://developer.nvidia.com/clara}} has enabled FL. It adopts a centralized architecture and encrypted communication channel. The targeted users of Clara FL is hospitals and medical institutions. Ping An Technology aims to build a federated learning system named Hive \cite{pinganhive}, which targets at the financial industries. While Clara FL provides APIs and documents, we cannot find the official documents of Hive.

\subsubsection{Summary}

\begin{table*}[]
\caption{The comparison among some existing FLSs. The notations used in this table are the same as Table~\ref{tbl:study_compare}. The cell is left empty if the system does not support the corresponding feature. There is no release version for FedML.}
\label{tbl:fls_compare}
\newcommand{\y}{\cmark}
\newcommand{\n}{\xmark}
\centering
\resizebox{\linewidth}{!}{%
\begin{tabular}{|c|c|c|c|c|c|c|}
\hline
\multicolumn{2}{|c|}{Supported features} 
            & FATE 1.5.0& TFF 0.17.0& PySyft 0.3.0& PaddleFL 1.1.0& FedML\\ \hline
\multirow{5}{*}{Operation systems} 
& Mac       & \y         & \y         & \y           & \y             & \y   \\ \cline{2-7} 
& Linux     & \y         & \y         & \y           & \y             & \y   \\ \cline{2-7} 
& Windows   &          &          & \y           & \y             & \y   \\ \cline{2-7}
& iOS       &          &          &            &              & \y   \\ \cline{2-7}
& Android   &          &          &            &              & \y   \\ \hline
\multirow{2}{*}{Data partitioning} 
& horizontal& \y         & \y         & \y           & \y             & \y   \\ \cline{2-7} 
& vertical  & \y         &          &            & \y             & \y   \\ \hline
\multirow{3}{*}{Models} 
& NN        & \y         & \y         & \y           & \y             & \y   \\ \cline{2-7} 
& DT        & \y         &         &            &              &   \\ \cline{2-7} 
& LM        & \y         & \y         & \y           & \y             & \y   \\ \hline
\multirow{2}{*}{Privacy Mechanisms} 
& DP        &         & \y         & \y           & \y             &\y   \\ \cline{2-7} 
& CM        & \y         &          & \y           & \y             &    \\ \hline
\multirow{2}{*}{Communication} 
& simulated & \y         & \y         & \y           & \y             & \y   \\ \cline{2-7} 
& distributed & \y       &          & \y           & \y             & \y   \\ \hline
\multirow{2}{*}{Hardwares} 
& CPUs      & \y         & \y         & \y           & \y             & \y   \\ \cline{2-7}
& GPUs      &         & \y         & \y           &            & \y   \\ \hline
\end{tabular}
}
\end{table*}

Overall, FATE, PaddleFL, and FedML try to provide algorithm-level APIs for users to use directly, while TFF and PySyft try to provide more detailed building blocks so that the developers can easily implement their FL process. Table \ref{tbl:fls_compare} shows the comparison between the open-source systems. In the algorithm level, FATE is the most comprehensive system that supports many machine learning models under both horizontal and vertical settings. TFF and PySyft only implement FedAvg, which is a basic framework in FL as shown in Section \ref{sec:ind_stu}. PaddleFL supports several horizontal FL algorithms currently on NNs and logistic regression. FedML integrates several state-of-the-art FL algorithms such as FedOpt \cite{reddi2020adaptive} and FedNova \cite{wang2020tackling}. Compared with FATE, TFF, and FedML, PySyft and PaddleFL provide more privacy mechanisms. PySyft covers all the listed features that TFF supports, while TFF is based on TensorFlow and PySyft works better on PyTorch. \rev{Based on the popularity on GitHub, PySyft is currently the most impactful federated learning system in the machine learning community.}

\section{System Design}
\label{sec:design}
\begin{figure}
\begin{center}
\includegraphics[width=0.8\columnwidth]{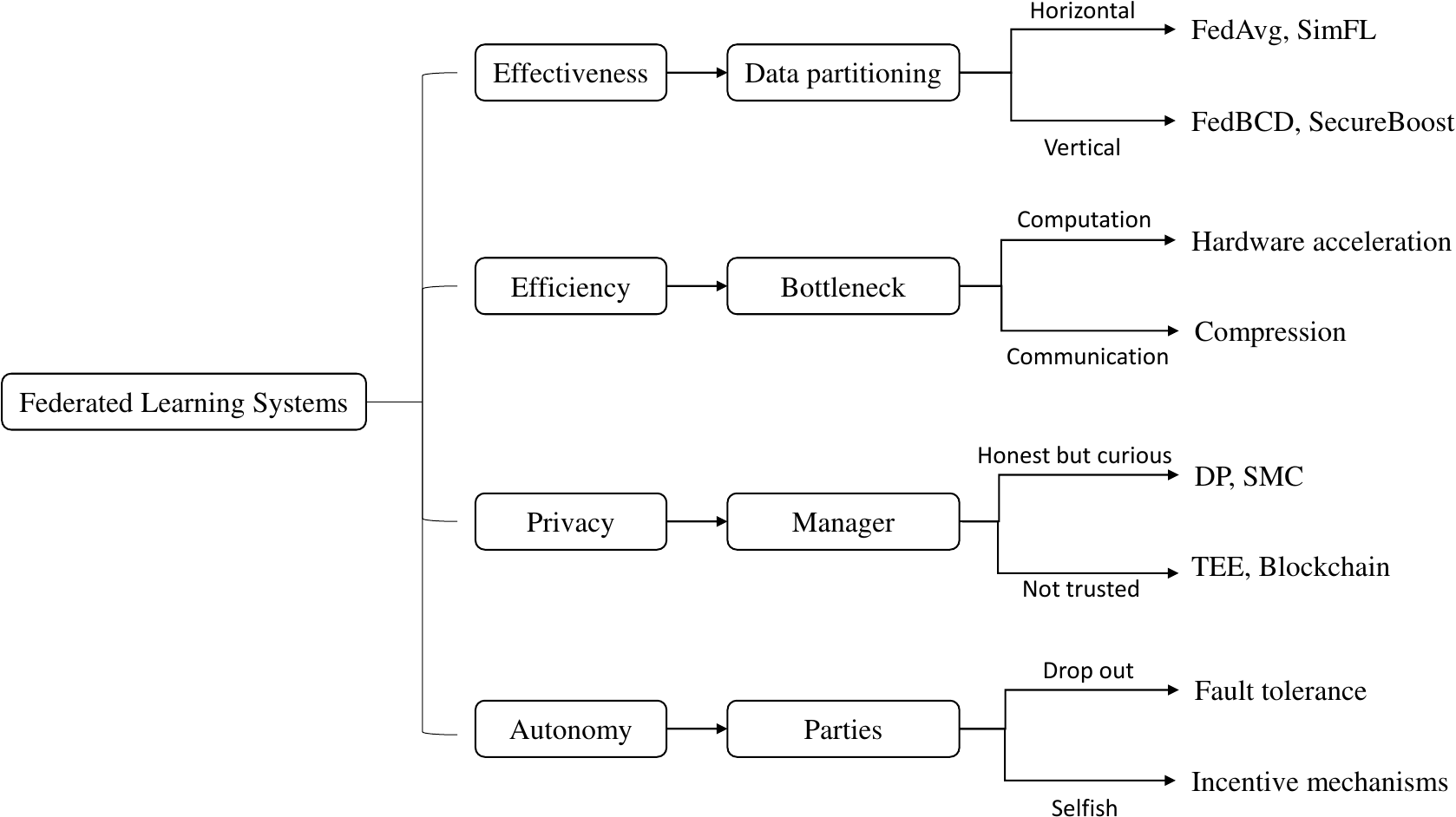}
\caption{The design factors of FLSs}
\label{fig:sys_design}
\end{center}
\end{figure}

Figure~\ref{fig:sys_design} shows the factors that need to be considered in the design of an FLS. \rev{Here effectiveness, efficiency, and privacy are three important metrics of FLSs, which are also main research directions of federated learning. Inspired by federated database \cite{sheth1990federated}, we also consider autonomy, which is necessary to make FLSs practical.} Next, we explain these factors in detail.

\subsection{Effectiveness}
The core of an FLS is an (multiple) effective algorithm (algorithms). To determine the algorithm to be implemented from lots of existing studies as shown in Table~\ref{tbl:study_compare}, we should first check the data partitioning of the parties. If the parties have the same features but different samples, one can use FedAvg \cite{mcmahan2016communication} for NNs and SimFL \cite{li2019practical} for trees. If the parties have the same sample space but different features, one can use FedBCD \cite{liu2019communication} for NNs and SecureBoost \cite{cheng2019secureboost} for trees.

\subsection{Privacy}
An important requirement of FLSs is to protect the user privacy. Here we analyze the reliability of the manager. If the manager is honest and not curious, then we do not need to adopt any additional technique, since the FL framework ensures that the raw data is not exchanged. If the manager is honest but curious, then we have to take possible inference attacks into consideration. The model parameters may also expose sensitive information about the training data. One can adopt differential privacy \cite{geyer2017differentially,choudhury2019differential,mcmahan2017learning} to inject random noises into the parameters or use SMC \cite{bonawitz2016practical,hardy2017private,bonawitz2017practical} to exchanged encrypted parameters. If the manager cannot be trusted at all, then we can use trusted execution environments \cite{chen2020training} to execute the code in the manager. Blockchain is also an option to play the role as a manager \cite{kim2018device}.

\subsection{Efficiency}
\rev{Efficiency is an important factor in the success of many existing systems such as XGBoost \cite{chen2016xgboost} and ThunderSVM \cite{wenthundersvm18}. Since federated learning involves multi-rounds training and communication, the computation and communication costs may be large, which increases the threshold of usage of FLSs.} To increase the efficiency, the most effective way is to deal with the bottleneck. If the bottleneck lies in the computation, we can use powerful hardware such as GPUs \cite{cook2012cuda} and TPUs \cite{jouppi2017datacenter}. If the bottleneck lies in the communication, the compression techniques \cite{bernstein2018signsgd,konevcny2016federated,sattler2019robust} can be applied to reduce the communication size. 

\subsection{Autonomy}
Like federated databases \cite{sheth1990federated}, a practical FLS has to consider the autonomy of the parties. The parties may drop out (e.g., network failure) during the FL process, especially in the cross-device setting where the scale is large and the parties are unreliable \cite{kairouz2019advances}. \rev{Thus, the FLS should be robust and stable, which can tolerate the failure of parties or reduce the number of failure cases. Google has developed a practical FLS \cite{bonawitz2019towards}. In their system, they monitor devices' health statistics to avoid wasting devices' battery or bandwidth. Also, the system will complete the current round or restart from the results of the previously committed round if there are failures. \citet{zhang2020blockchain} propose a blockchain-based approach to detect the device disconnection. Robust secure aggregation \cite{bell2020secure} is applicable to protect the communicated message in case of party drop out.} Besides the disconnection issues, the parties may be selfish and are not willing to share the model with good quality. Incentive mechanisms \cite{kang2019incentive,kang2019incentive2} can encourage the participation of the parties and improve the final model quality. 

\subsection{The Design Reference}
\rev{Based on our taxonomy shown in Section \ref{sec:taxonomy} and the design factors shown in Figure \ref{fig:sys_design}, we derive a simple design reference for developing an FLS.}

\rev{The first step is to identity the participated entities and the task, which significantly influence the system design. The participated entities determines the communication architecture, the data partitioning and the scale of federation. The task determines the suitable machine learning models to train. Then, we can choose or design a suitable FL algorithm according to the above attributes and Table \ref{tbl:study_compare}. After fixing the FL algorithm, to satisfy the privacy requirements, we may determine the privacy mechanisms to protect the communicated messages. DP is preferred if efficiency is more important than model performance compared with SMC. Last, incentive mechanism can be considered to enhance the system. Existing systems \cite{he2020fedml,bonawitz2019towards} usually do not support incentive mechanisms. However, incentive mechanisms can encourage the parties to participate and contribute in the system and make the system more attractive. Shapley value \cite{wang2020principled,wang2019measure} is a fair approach that can be considered.}

For real-world applications of federated learning systems, please refer to Section 4 of the supplementary material.

\subsection{\rev{Evaluation}}
\label{sec:sys_eval}

\rev{The evaluation of FLSs is very challenging. According to our studied system factors, it has to cover the following aspects: (1) model performance, (2) system security, (3) system efficiency, and (4) system robustness.}

\rev{For the evaluation of the model, there are two different settings. One is to evaluate the performance (e.g., prediction accuracy) of the final global model on a global dataset. The other one is to evaluate the performance of the final local models on the corresponding non-IID local datasets. The evaluation setting depends on the objective of FL, i.e., learn a global model or learn personalized local models.}

\rev{While theoretical security/privacy guarantee is a good evaluation metric for system security, another way is to conduct membership inference attacks \cite{shokri2017membership} or model inversion attacks \cite{fredrikson2015model} to test the system security. These attacks can be conducted in two ways: (1) white-box attack: the attacker has access to all the exchanged models during the FL process. (2) black-box attack: the attacker only has access to the final output model. The attack success ratio can be an evaluation metric for the system security.
}

\rev{The efficiency of the system includes two parts: computation efficiency and communication efficiency. An intuitive metric is the training time, including the computation and communication time. Note that FL is usually a multi-round process. Thus, for a fair comparison, one approach is to use time per round as a metric. Another approach is to record the time or round to achieve the same target performance \cite{karimireddy2020scaffold,li2021model}.
}

\rev{It is challenging to quantifying the robustness of an FLS. A possible solution is to use a similar metric as robust secure aggregation, i.e., the maximum number of disconnected parties that can tolerate during the FL process. 
}
\section{Case Study}
\label{sec:case_study}
In this section, we present several real-world applications of FL according to our taxonomy, as summarized in Table \ref{tbl:case-req}.

\begin{table*}[]
\small
    \centering
    \caption{Requirements of the real-world federated systems}
    \label{tbl:case-req}
    \begin{tabular}{|c|c|c|c|}
    \hline
        System Aspect & Mobile Service & Healthcare & Financial \\\hline
        Data Partitioning & Horizontal Partitioning & Hybrid Partitioning & Vertical Partitioning\\\hline
        Machine Learning Model & No specific Models & No specific Models & No specific Models\\\hline
        Scale of Federations & Cross-device & Cross-silo & Cross-silo\\\hline
        Communication Architecture & Centralized &  Distributed & Distributed\\\hline
        Privacy Mechanism & DP & DP/SMC & DP/SMC\\\hline
        Motivation of Federation & Incentive Motivated &  Policy Motivated & Interest Motivated \\ \hline
    \end{tabular}
\end{table*}

\subsection{Mobile Service}
There are many corporations providing predicting service to their mobile users, such as Google Keyboard~\cite{yang2018applied}, Apple's emoji suggestion and QuickType~\cite{adp2017learning}. These services bring much convenience to the users. However, the training data come from users' edge devices, like smartphones. If the company collects data from all users and trains a global model, it might potentially cause privacy leakage. On the other hand, the data of each single user are insufficient to train an accurate prediction model. FL enables these companies to train an accuracy prediction model without accessing users' original data, which means protecting users' privacy. In the framework of FLSs, the users calculate and send their local models instead of their original data. That means a Google Keyboard user can enjoy an accurate prediction for the next word while not sharing his/her input history. If FLS can be widely applied to such prediction services, there will be much less data leakage since data are always stored in the edge. 

In such a scenario, data are usually horizontally split into millions of devices. Hence, the limitation of single device computational resource and the bandwidth are two major problems. Besides, the robustness of the system should also be considered since a user could join or leave the system at anytime. In other words, a centralized, cross-device FLS on horizontal data should be designed for such prediction services.

Although the basic framework of an FLS can have somehow protected individuals' privacy, it may not be secure against inference attacks~\cite{shokri2017membership}. Some additional privacy mechanisms like differential privacy should be leveraged to ensure the indistinguishability of individuals. Here secure multi-party computation may not be appropriate since each device has a weak computation capacity and cannot afford expensive encryption operations. Apart from guaranteeing users' privacy, some incentive mechanisms should be developed to encourage users to contribute their data. In reality, these incentives could be vouchers or additional service. 

\subsection{Healthcare}
Modern health systems require cooperation among research institutes, hospitals, and federal agencies to improve health care of the nation~\cite{friedman2010achieving}. Moreover, collaborative research among countries is vital when facing global health emergencies, like COVID-19~\cite{covid19who}. These health systems mostly aim to train a model for the diagnosis of a disease. These models for diagnosis should be as accurate as possible. However, the information of patients are not allowed to transfer under some regulations such as GDPR~\cite{albrecht2016gdpr}. The privacy of data is even more concerned in international collaboration. Without solving the privacy issue, the collaborative research could be stagnated, threatening the public health. The data privacy in such collaboration is largely based on confidentiality agreement. However, this solution is based on ``trust'', which is not reliable. FL makes the cooperation possible because it can ensure the privacy theoretically, which is provable and reliable. In this way, every hospital or institute only has to share local models to get an accurate model for diagnosis.

In such a scenario, the health care data is partitioned both horizontally and vertically: each party contains health data of residents for a specific purpose (e.g., patient treatment), but the features used in each party are diverse. The number of parties is limited and each party usually has plenty of computational resource. In other words, a private FLS on hybrid partitioned data is required. One of the most challenging problems is how to train the hybrid partitioned data. The design of the FLS could be more complicated than a simple horizontal system. In a federation of healthcare, there is probably no central server. So, another challenging part is the design of a decentralized FLS, which should also be robust against some dishonest or malicious parties. Moreover, the privacy concern can be solved by additional mechanisms like secure multi-party computation and differential privacy. The collaboration is largely motivated by regulations.

\subsection{Finance}
A federation of financial consists of banks, insurance companies, etc. They often hope to cooperate in daily financial operations. For example, some `bad' users might pack back a loan in one back with the money borrowed from another bank. All the banks want to avoid such malicious behavior while not revealing other customers' information. Also, insurance companies also want to learn from the banks about the reputation of customers. However, a leakage of `good' customers' information may cause loss of interest or some legal issues.

This kind of cooperation can happen if we have a trusted third party, like the government. However, in many cases, the government is not involved in the federation or the government is not always trusted. So, an FLS with privacy mechanisms can be introduced. In the FLS, the privacy of each bank can be guaranteed by theoretical proved privacy mechanisms.

In such a scenario, financial data are often vertically partitioned, linked by user ID. Training a classifier in vertically partitioned data is quite challenging. Generally, the training process can be divided into two parts: privacy-preserving record linkage~\cite{vatsalan2017privacy} and vertical federated training. The first part aims to find links between vertical partitioned data, and it has been well studied. The second part aims to train the linked data without sharing the original data of each party, which still remains a challenge. The cross-silo and decentralized setting are applied in this federation. Also, some privacy mechanisms should be adopted in this scenario and the participant can be motivated by interest.

\section{Vision}
\label{sec:vision}
\rev{In this section, we show interesting directions to work on in the future. Although some directions are already covered in existing studies introduced in Section 4, we believe they are important and provide more insights on them.}

\subsection{Heterogeneity}
The heterogeneity of the parties is an important characteristic in FLSs. Basically, the parties can differ in the accessibility, privacy requirements, contribution to the federation, and reliability. Thus, it is important to consider such practical issues in FLSs.

\noindent \textbf{Dynamic scheduling}
Due to the instability of the parties, the number of parties may not be fixed during the learning process. However, the number of parties is fixed in many existing studies and they do not consider the situations where there are entries of new parties or departures of the current parties. The system should support dynamic scheduling and have the ability to adjust its strategy when there is a change in the number of parties. There are some studies addressing this issue. For example, Google's  system~\cite{bonawitz2019towards} can tolerate the drop-out of the devices. Also, the emergence of blockchain~\cite{zheng2018blockchain} can be an ideal and transparent platform for multi-party learning. \rev{However, to the best of our knowledge, there is no work that study a increasing number of parties during FL. In such a case, more attention may be paid to the later parties, as the current global model may have been welled trained on existing parties.} 

\noindent \textbf{Diverse privacy restrictions}
Little work has considered the privacy heterogeneity of FLSs, where the parties have different privacy requirements. The existing systems adopt techniques to protect the model parameters or gradients for all the parties on the same level. However, the privacy restrictions of the parties usually differ in reality. It would be interesting to design an FLS which treats the parties differently according to their privacy restrictions. The learned model should have a better performance if we can maximize the utilization of data of each party while not violating their privacy restrictions. \rev{The heterogeneous differential privacy~\cite{alaggan2015heterogeneous} may be useful in such settings, where users have different privacy attitudes and expectations.}

% \subsubsection{Intelligent benefits}
\noindent \textbf{Intelligent benefits}

Intuitively, one party should gain more from the FLS if it contributes more information. \rev{Existing incentive mechanisms are mostly based on Shapley values \cite{wang2020principled,wang2019measure}, the computation overhead is a major concern. A computation-efficient and fair incentive mechanism needs to be developed.}

% \subsubsection{Robustness}

\subsection{System Development}
To boost the development of FLSs, besides the detailed algorithm design, we need to study from a high-level view.

\noindent \textbf{System architecture}
Like the parameter server \cite{ho2013more} in deep learning which controls the parameter synchronization, some common system architectures are needed to be investigated for FL. Although FedAvg is a widely used framework, the applicable scenarios are still limited. \rev{For example, while unsupervised learning \cite{mcmahan2016communication,li2021model,li2018federated} still adopt model averaging as the model aggregation method, which cannot work if the parties want to train heterogeneous models.} We want a general system architecture, which provides many aggregation methods and learning algorithms for different settings.

\noindent \textbf{Model market}
\rev{Model market \cite{vartak2016modeldb} is a promising platform for model storing, sharing, and selling. An interesting idea is to use the model market for federated learning. }The party can buy the models to conduct model aggregation locally. Moreover, it can contribute its models to the market with additional information such as the target task. Such a design introduces more flexibility to the federation and is more acceptable for the organizations, since the FL just like several transactions. A well evaluation of the models is important in such systems. The incentive mechanisms may be helpful \cite{weng2019deepchain,kang2019incentive,kang2019incentive2}.

\noindent \textbf{Benchmark}
As more FLSs are being developed, a benchmark with representative data sets and workloads is quite important to evaluate the existing systems and direct future development. Although there have been quite a few benchmarks \cite{caldas2018leaf,hu2020oarf,he2020fedml}, no benchmark has been widely used in the experiments of federated learning studies. We need a robust benchmark which has representative datasets and strict privacy evaluation. \rev{Also, comprehensive evaluation metric including model performance, system efficiency, system security, and system robustness is often ignored in existing benchmarks. The evaluation of model performance on non-IID datasets and system security under data pollution needs more investigation.}

\noindent \textbf{Data life cycles}
\HBS{Learning is simply one aspects of a federated system. A data life cycle \cite {polyzotis2018data} consists of multiple stages including data creation, storage, use, share, archive and destroy.  For the data security and privacy of the entire application, we need to invent new data life cycles under FL context. Although data sharing is clearly one of the focused stage, the design of FLSs also affects other stages. For example, data creation may help to prepare the data and features that are suitable for FL.}

\subsection{FL in Domains}

\noindent \textbf{Internet-of-thing}
Security and privacy issues have been a hot research area in fog computing and edge computing, due to the increasing deployment of Internet-of-thing applications. For more details, readers can refer to some recent surveys~\cite{Stojmenovic:2016:OFC:3072121.3072138,Yi2015SecurityAP,8026115}. FL can be one potential approach in addressing the data privacy issues, while still offering reasonably good machine learning models~\cite{lim2019federated,nguyen2018dot}. The additional key challenges come from the computation and energy constraints. The mechanisms of privacy and security introduces runtime overhead. For example, \citet{Jiang:2019:LPC:3302505.3310070} apply independent Gaussian random projection to improve the data privacy, and then the training of a deep network can be too costly. The authors need to develop a new resource scheduling algorithm to move the workload to the nodes with more computation power. Similar issues happen in other environments such as vehicle-to-vehicle networks~\cite{samarakoon2018federated}.

% \subsubsection{Regulations}
\noindent \textbf{Regulations}
While FL enables collaborative learning without exposing the raw data, it is still not clear how FL comply with the existing regulations. For example, GDPR proposes limitations on data transfer. Since the model and gradients are actually not safe enough, is such limitation still apply to the model or gradients? Also, the ``right to explainability'' is hard to execute since the global model is an averaging of the local models. The explainability of the FL models is an open problem \cite{gunning2017explainable,samek2017explainable}. Moreover, if a user wants to delete its data, should the global model be retrained without the data \cite{ginart2019making}? There is still a gap between the FL techniques and the regulations in reality. We may expect the cooperation between the computer science community and the law community.

\section{Conclusion}
\label{sec:conc}
Many efforts have been devoted to developing federated learning systems (FLSs). A complete overview and summary for existing FLSs is important and meaningful. Inspired by the previous federated systems, we have shown that heterogeneity and autonomy are two important factors in the design of practical FLSs. Moreover, with six different aspects, we provide a comprehensive categorization for FLSs. Based on these aspects, we also present the comparison on features and designs among existing FLSs. More importantly, we have pointed out a number of opportunities, ranging from more benchmarks to integration of emerging platforms such as blockchain. \HBS{FLSs will be an exciting research direction, which calls for the effort from machine learning, system and data privacy communities.}

\section*{Acknowledgement}
This work is supported by a MoE AcRF Tier 1 grant (T1 251RES1824), an SenseTime Young Scholars Research Fund, and a MOE Tier 2 grant (MOE2017-T2-1-122) in Singapore.

\bibliographystyle{plainnat}
\bibliography{ref}

\end{document}